\documentclass[letterpaper, 10 pt, conference]{ieeeconf}  % Comment this line out if you need a4paper

\IEEEoverridecommandlockouts                              % This command is only needed if 
\usepackage{amsmath}
\usepackage{amssymb}
\usepackage{wrapfig}
\usepackage{subfig}
\usepackage{caption}
\usepackage{color}
\usepackage{algorithm}
\usepackage{algorithmic}
\usepackage{graphicx}

\usepackage{pgfplots}
\usepackage{tikz}
\usepackage{tikz-3dplot}

\usepackage{ragged2e}
\usepackage{fp}

\usepackage{ifthen}

\usepackage{tikz}
\usetikzlibrary{shapes.geometric, arrows, calc,automata,positioning,decorations,decorations.text}

\usetikzlibrary{mindmap,snakes}

\usepackage{multirow}

\usepackage{algorithm}
\usepackage{algorithmic}
\usepackage{graphicx}
\usepackage[english]{babel}
\usepackage{array}
\usepackage[export]{adjustbox}

\graphicspath{{images/}}

\usepackage{xcolor}

\usepackage[colorlinks=true,
    linkcolor=black,
    linkbordercolor=white,
    urlcolor=red!90!blue,
    urlbordercolor=white,
    ]{hyperref}

\usepackage{colortbl}
\usepackage{makecell}
\usepackage{pifont}

\newcommand{\cmark}{\ding{51}}%
\newcommand{\xmark}{\ding{55}}

\title{\LARGE \bf
Towards Deep Learning Assisted Autonomous UAVs for Manipulation Tasks in GPS-Denied Environments
}

\author{Ashish~Kumar$^{\dagger}$, Mohit Vohra$^{\dagger}$, Ravi Prakash$^{\dagger}$, L. Behera$^{\dagger}$, \textit{Senior Member IEEE} \\
{\tt\small{\{\url{ https://github.com/ashishkumar822}}, \url{https://youtu.be/kxg9xmr3aEM}\}}
\thanks{$^{\dagger}$All authors are with the Department of Electrical Engineering, Indian Institute of Technology, Kanpur
        {\tt\small \{krashish,mvohra,ravipr,lbehera\}@iitk.ac.in}}%
}

\begin{document}

\maketitle
\thispagestyle{empty}
\pagestyle{empty}

%%%%%%%%%%%%%%%%%%%%%%%%%%%%%%%%%%%%%%%%%%%%%%%%%%%%%%%%%%%%%%%%%%%%%%%%%%%%%%%%
\begin{justify}

\begin{abstract}
In this work, we present a pragmatic approach to enable unmanned aerial vehicle (UAVs) to autonomously perform highly complicated tasks of object pick and place. This paper is largely inspired by challenge-$2$ of MBZIRC $2020$ and is primarily focused on the task of assembling large $3$D structures in outdoors and GPS-denied environments. Primary contributions of this system are: (\textit{\color{black}i}) a novel computationally efficient deep learning based unified multi-task visual perception system for target localization, part segmentation, and tracking, (\textit{\color{black}ii}) a novel deep learning based grasp state estimation, (\textit{\color{black}iii}) a retracting electromagnetic gripper design, (\textit{\color{black}iv}) a remote computing approach which exploits state-of-the-art MIMO based high speed ($5000$Mb/s) wireless links to allow the UAVs to execute compute intensive tasks on remote high end compute servers, and (\textit{\color{black}v}) system integration in which several system components are weaved together in order to develop an optimized software stack. We use DJI Matrice-$600$ Pro, a hex-rotor UAV and interface it with the custom designed gripper. Our framework is deployed on the specified UAV in order to report the performance analysis of the individual modules. Apart from the manipulation system, we also highlight several hidden challenges associated with the UAVs in this context.

\end{abstract}

\end{justify}

%%%%%%%%%%%%%%%%%%%%%%%%%%%%%%%%%%%%%%%%%%%%%%%%%%%%%%%%%%%%%%%%%%%%%%%%%%%%%%%%
%

\section{Introduction}
\label{sec_intro}
%Object manipulation is one of the fundamental tasks which is mastered by a humans in their early phases of life. This task, later on, becomes so important that several high level tasks such as sorting a heap of items, doing dishes, screw fastening etc. can only be accomplished by performing object manipulation in a sense or another. 
Despite being an easy task for humans, object manipulation using robots (robotic manipulation) is not very straight forward. Robotic manipulation using multi-DoF robotic arms has been studied extensively in the literature and has witnessed major breakthrough in the past decade, thanks to the advent of deep learning based visual perception methods and high performance parallel computing hardware (GPUs, TPUs). Various international level robotics challenges such as DARPA, Amazon Picking $2016$ and Robotics Challenge-$2017$, have played a major role in pushing state-of-the-art in this area.
\par
Autonomous robotic manipulation using UAVs, on the other hand, has just started marking its presence. It is because, developing low cost UAVs or vertical-takeoff-landing (VTOL) micro-aerial-vehicles (MAV)\footnote{used interchangeably with UAV throughout the paper} has only recently become possible. The low cost revolution is primarily driven by the drone manufacturers such as DJI which covers  $\sim70\%$ of the drone market worldwide and provides low cost industrial drones for manual operation. Apart from that, the open-source autopilot projects such as Ardupilot \cite{ardupilot} have also contributed in this revolution. The above reasons have resulted in their increased popularity among the worldwide researches to develop algorithms for their autonomous operations. In the area of manipulation using UAVs, one of the most prominent and visible effort comes from E-commerce giant Amazon, a multi-rotor UAV based delivery system. 

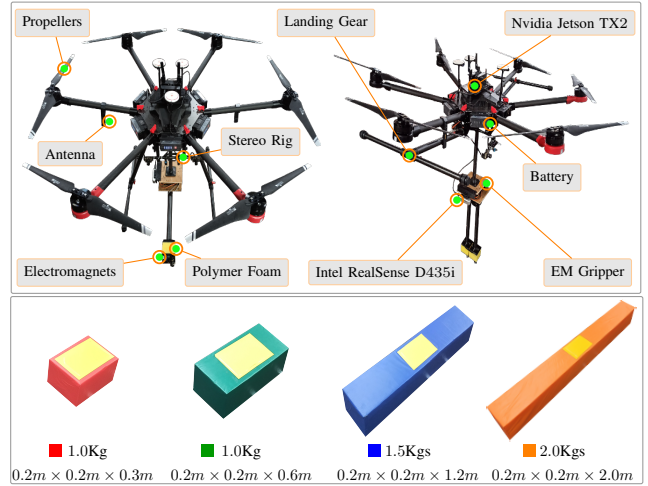
\begin{figure}[t]
    \centering
\begin{tikzpicture}

\node (m600) [scale=1, xshift=0ex]
{
\tikz{
\node (m600_w_retracted_gripper) [xshift = -13ex]{\includegraphics[width=25ex,height=20ex]{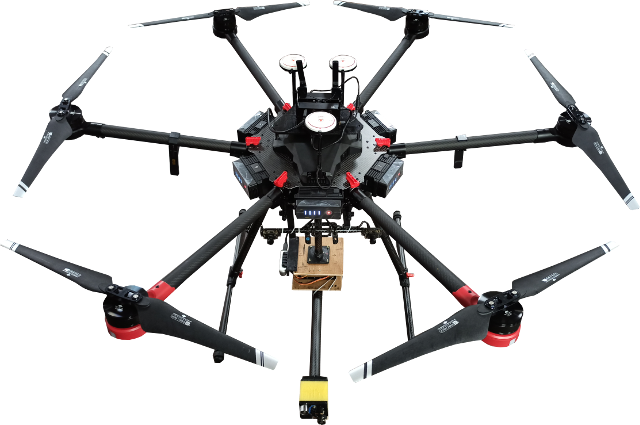}};
\node (m600_w_extended_gripper) [xshift = 13ex]{\includegraphics[width=25ex,height=20ex]{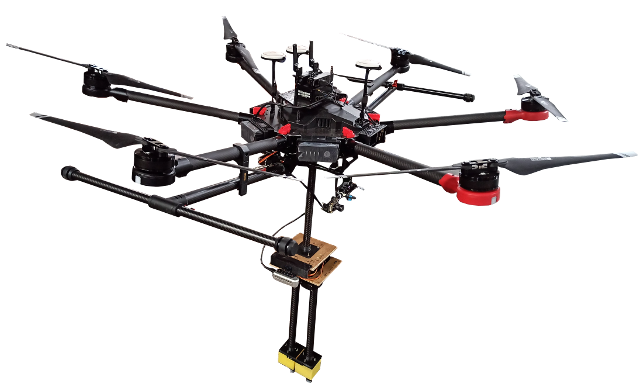}};
\node (m600_boundary) [draw=black!40!white,rounded corners=0.25mm,rectangle, align=center,minimum width=54ex,minimum height=25ex,scale=0.98]{};
\FPeval{\circwidth}{0.15}
\FPeval{\dotscale}{0.3}
\FPeval{\circscale}{0.5}
\colorlet{dotcolor}{white!10!green}
\colorlet{circcolor}{orange}
\colorlet{linkcolor}{orange}
\colorlet{boxdrawcolor}{white!60!orange}
\colorlet{boxfillcolor}{white!90!black}
\FPeval{\txtscale}{0.8}
\node (batterydot) [fill=dotcolor, circle, xshift =13.8ex,yshift=2.0ex, scale=\dotscale]{};
\node (batterydotcirc) [draw=circcolor, circle, xshift =13.8ex,yshift=2.0ex, line width=\circwidth ex,scale=\circscale]{};
\node (battery) [draw=boxdrawcolor,fill=boxfillcolor,rounded corners=0.3mm, rectangle,align=center,xshift =19ex,yshift=-2.5ex,scale=\txtscale]{\scriptsize Battery};
\draw [-, linkcolor] (batterydotcirc) -- (battery);
\node (antennadot) [fill=dotcolor, circle, xshift =-18.2ex,yshift=2.2ex, scale=\dotscale]{};
\node (antennadotcirc) [draw=circcolor, circle, xshift =-18.2ex,yshift=2.2ex, line width=\circwidth ex,scale=\circscale]{};
\node (antenna) [draw=boxdrawcolor,fill=boxfillcolor,rounded corners=0.3mm, rectangle,align=center,xshift =-21.5ex,yshift=-0.5ex,scale=\txtscale]{\scriptsize Antenna};
\draw [-, linkcolor] (antennadotcirc) -- (antenna);
\node (stereorigdot) [fill=dotcolor, circle, xshift =-12.0ex,yshift=-0.8ex, scale=\dotscale]{};
\node (stereorigdotcirc) [draw=circcolor, circle, xshift =-12.0ex,yshift=-0.8ex, line width=\circwidth ex,scale=\circscale]{};
\node (stereorig) [draw=boxdrawcolor,fill=boxfillcolor,rounded corners=0.3mm, rectangle,align=center,xshift =-5.5ex,yshift=0.5ex,scale=\txtscale]{\scriptsize Stereo Rig};
\draw [-, linkcolor] (stereorigdotcirc) -- (stereorig);
\node (propsdot) [fill=dotcolor, circle, xshift =-22.0ex,yshift=6.65ex, scale=\dotscale]{};
\node (propsdotcirc) [draw=circcolor, circle, xshift =-22.0ex,yshift=6.65ex, line width=\circwidth ex,scale=\circscale]{};
\node (props) [draw=boxdrawcolor,fill=boxfillcolor,rounded corners=0.3mm, rectangle,align=center,xshift =-23ex,yshift=10.5ex,scale=\txtscale]{\scriptsize Propellers};
\draw [-, linkcolor] (propsdotcirc) -- (props);
\node (gripperdot) [fill=dotcolor, circle, xshift =13.5ex,yshift=-3.0ex, scale=\dotscale]{};
\node (gripperdotcirc) [draw=circcolor, circle, xshift =13.5ex,yshift=-3.0ex, line width=\circwidth ex,scale=\circscale]{};
\node (gripper) [draw=boxdrawcolor,fill=boxfillcolor,rounded corners=0.3mm, rectangle,align=center,xshift =22ex,yshift=-10.5ex,scale=\txtscale]{\scriptsize EM Gripper};
\draw [-, linkcolor] (gripperdotcirc) -- (gripper);
\node (foamdot) [fill=dotcolor, circle, xshift =-12.6ex,yshift=-8.5ex, scale=\dotscale]{};
\node (foamdotcirc) [draw=circcolor, circle, xshift =-12.6ex,yshift=-8.5ex, line width=\circwidth ex,scale=\circscale]{};
\node (foam) [draw=boxdrawcolor,fill=boxfillcolor,rounded corners=0.3mm, rectangle,align=center,xshift =-7.5ex,yshift=-10.5ex,scale=\txtscale]{\scriptsize Polymer Foam};
\draw [-, linkcolor] (foamdotcirc) -- (foam);
\node (emagsdot) [fill=dotcolor, circle, xshift =-14.0ex,yshift=-9.2ex, scale=\dotscale]{};
\node (emagsdotcirc) [draw=circcolor, circle, xshift =-14.0ex,yshift=-9.2ex, line width=\circwidth ex,scale=\circscale]{};
\node (emags) [draw=boxdrawcolor,fill=boxfillcolor,rounded corners=0.3mm, rectangle,align=center,xshift =-21.5ex,yshift=-10.5ex,scale=\txtscale]{\scriptsize Electromagnets};
\draw [-, linkcolor] (emagsdotcirc) -- (emags);
\node (computerdot) [fill=dotcolor, circle, xshift =12.5ex,yshift=5.2ex, scale=\dotscale]{};
\node (computerdotcirc) [draw=circcolor, circle, xshift =12.5ex,yshift=5.2ex, line width=\circwidth ex,scale=\circscale]{};
\node (computer) [draw=boxdrawcolor,fill=boxfillcolor,rounded corners=0.3mm, rectangle,align=center,xshift =20.5ex,yshift=10.5ex,scale=\txtscale]{\scriptsize Nvidia Jetson TX$2$};
\draw [-, linkcolor] (computerdotcirc) -- (computer);
\node (realsensedot) [fill=dotcolor, circle, xshift =11.0ex,yshift=-4.4ex, scale=\dotscale]{};
\node (reaslsensedotcirc) [draw=circcolor, circle, xshift =11.0ex,yshift=-4.4ex, line width=\circwidth ex,scale=\circscale]{};
\node (realsense) [draw=boxdrawcolor,fill=boxfillcolor,rounded corners=0.3mm, rectangle,align=center,xshift =5ex,yshift=-10.5ex,scale=\txtscale]{\scriptsize Intel RealSense D$435$i};
\draw [-, linkcolor] (reaslsensedotcirc) -- (realsense);
\node (landinggeardot) [fill=dotcolor, circle, xshift =7.0ex,yshift=-0.65ex, scale=\dotscale]{};
\node (landinggeardotcirc) [draw=circcolor, circle, xshift =7.0ex,yshift=-0.65ex, line width=\circwidth ex,scale=\circscale]{};
\node (landinggear) [draw=boxdrawcolor,fill=boxfillcolor,rounded corners=0.3mm, rectangle,align=center,xshift =0ex,yshift=10.5ex,scale=\txtscale]{\scriptsize Landing Gear};
\draw [-, linkcolor] (landinggeardotcirc) -- (landinggear);
}};
\node(bricks) [xshift = 0.5ex, yshift = -20.2 ex ]{\tikz{
\node (bricks_boundary) [draw=black!40!white,rounded corners=0.25mm,rectangle, align=center,minimum width=54ex,minimum height=16ex,yshift=-11.6ex, xshift=-4.5ex,scale=0.984]{};
\node (red) [xshift=-25ex, yshift=-10ex, scale = 0.6]{\includegraphics[width=10ex,height=10ex]{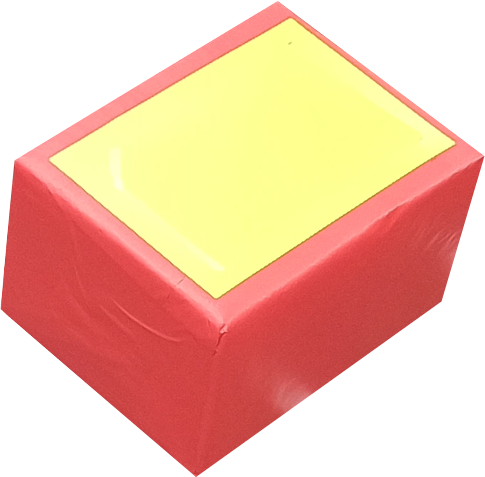}};
\node (green) [right of = red,  xshift=7ex, yshift=0ex, scale = 0.8 ] { \includegraphics[width=10ex,height=10ex]{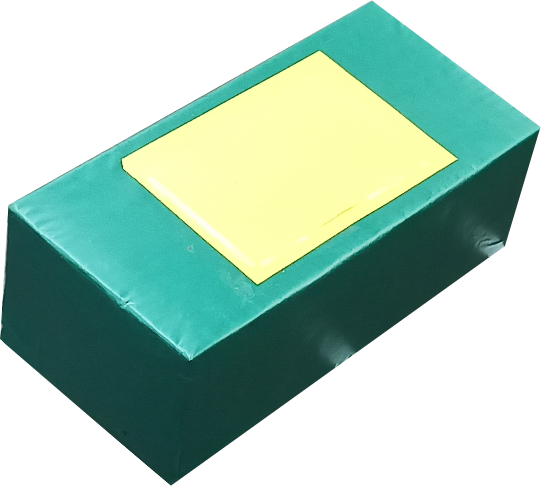}};
\node (blue) [right of = red,  xshift=21ex, yshift=0ex,scale=1.1]{\includegraphics[width=10ex,height=10ex]{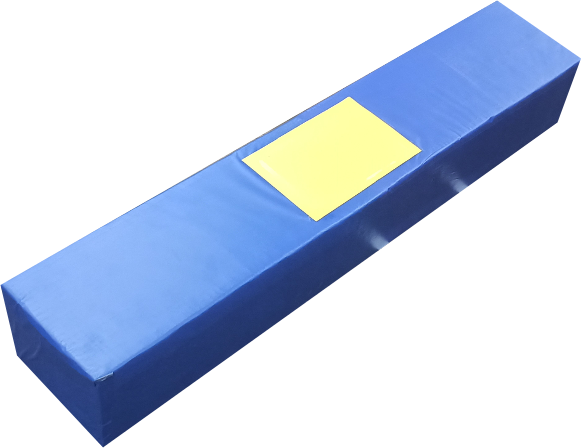}};
\node (orange) [right of = red, xshift=34ex, yshift=-0.0ex, rotate=5,scale =1.1]{\includegraphics[width=12ex,height=10ex]{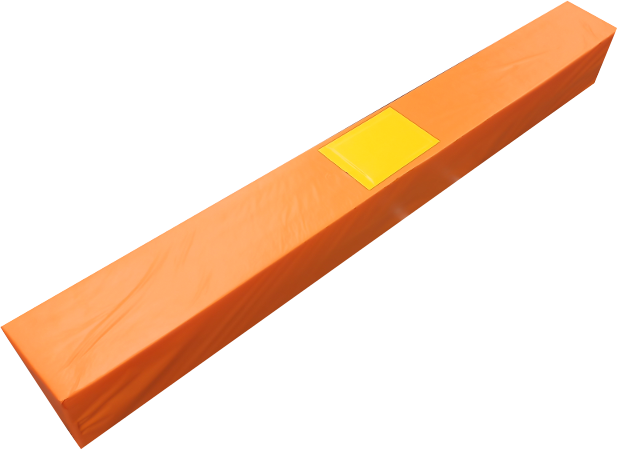}};
\node (red_weight) [below of = red , xshift=0.4ex, yshift=-0.5ex, scale = 0.6]{$1.0$Kg};
\node (red_mark) [left of = red_weight , fill=red, xshift=3.6ex, yshift=0.15ex, scale = 0.7]{};
\node (red_dim) [below of = red , xshift=0ex, yshift=-2.5ex, scale = 0.6]{$0.2m\times0.2m\times0.3m$};
\node (green_weight) [below of = green , xshift=0ex, yshift=-0.5ex, scale = 0.6]{$1.0$Kg};
\node (green_mark) [left of = green_weight , fill=green!60!black, xshift=3.5ex, yshift=0.15ex, scale = 0.7]{};
\node (green_dim) [below of = green , xshift=0ex, yshift=-2.5ex, scale = 0.6]{$0.2m\times0.2m\times0.6m$};
\node (blue_weight) [below of = blue , xshift=0ex, yshift=-0.5ex, scale = 0.6]{$1.5$Kgs};
\node (blue_mark) [left of = blue_weight , fill=blue, xshift=3.5ex, yshift=0.15ex, scale = 0.7]{};
\node (blue_dim) [below of = blue , xshift=0ex, yshift=-2.5ex, scale = 0.6]{$0.2m\times0.2m\times1.2m$};
\node (orange_weight) [below of = orange , xshift=0ex, yshift=-0.5ex, scale = 0.6]{$2.0$Kgs};
\node (orange_mark) [left of = orange_weight , fill=orange, xshift=3.5ex, yshift=0.15ex, scale = 0.7]{};\node (orange_dim) [below of = orange , xshift=0ex, yshift=-2.5ex, scale = 0.6]{$0.2m\times0.2m\times2.0m$};
}
};
\end{tikzpicture}
\caption{Top: M$600$ Pro with our gripper. Bottom: Bricks}
\label{fig_m600}
\end{figure}
\par
Another famous example is Mohammad Bin Zayed International Robotics Challenge (MBZIRC) $2020$, which is establishing new benchmarks for the drone based autonomy. The Challenge-$2$ of MBZIRC $2020$ requires a team of UAVs and an Unmanned Ground Vehicle (UGV) to locate, pick, transport and assemble fairly large cuboidal shaped objects (bricks (Fig.\ref{fig_m600})) into a given pattern to unveil tall $3$D structures. The bricks consist of identifiable ferromagnetic regions with yellow shade. 
\par
Inspired by the challenge discussed above, in this paper, we present an end-to-end industry grade solution for UAV based maniulation tasks. The developed solution is not limited to constrained workspaces and can be readily deployed into real world applications. Before diving into the actual solution, we first uncover several key challenges associated with UAV based manipulation below.
\subsection{Autonomous Control}
Multi-rotor VTOL MAVs UAVs are often termed as under-actuated, highly non-linear and complex dynamical system. These characteristics allow them to enjoy high agility and ability to perform complex maneuvering tasks such as mid-air flipping, sudden sharp turns etc. However, the agility comes at a cost which is directly related to its highly coupled control variables. Due to this reason, UAV control design in not an easy task. Hierarchical Proportional-Integral-Derivative (PID) control is one of the popular approaches which is generally employed for this purpose. Being quite simple, PID controllers are not intelligent enough to account for dynamic changes in the surroundings while their autonomous operations. To cope up with this, recently, the focus of researchers community has shifted towards machine learning based techniques for UAV control. However, due to data dependency and lack of generalization, learning based algorithms are still far from real deployable solution. Therefore, robust control of UAVs for autonomous operations still remains a challenge.
\subsection{Rotor Draft}
UAV rotor draft is a crucial factor which must be considered while performing drone based object manipulation. In order to execute a grasp operation, either the UAV must fly close to the target object or must have a long enough manipulator so that rotor draft neither disturbs the object nor the UAV itself. The former is only the case which is feasible but it's not an easy task. When flying at low altitudes, the rotor draft severely disturbs the UAV and therefore, poses a significant difficulty in front of stabilization and hovering algorithms. The latter, on the other hand, is not even possible as it would require a gripper of several meters long in order to diminish the effects of rotor draft. In addition, even a grasping mechanism $1-2m$ long, will increase the payload of UAV, resulting in quicker power source drainage. Also, such a long gripper will be infeasible to be accommodated into the UAV body.
\subsection{Dynamic Payload}
\label{subsec_dynpayload}
%\vspace{-2ex}
%
Dynamic payload attachment to the UAV is another important issue which arises after a successfull grasp operation. When a relatively heavier payload is attached to the UAV dynamically, the system characteristics and static thrust a.k.a hovering thrust also needs to be adjusted.  However, as static thrust is increased in order to compensate for the gravity, the issues of heating of batteries and propulsion system arises. This raises safety concerns about the battery system and also decreases operational flight time. Although, addition of small weights to the UAV can be ignored, this case becomes severe for larger payload weights i.e. when the weight to be attached is order of $4$ Kg while having an UAV with payload capacity $6$Kg. Thus, requirement of intelligent control algorithms which can handle non-linearities associated with UAVs, becomes evident. 
\subsection{Visual Perception}
In order to perform a seemingly easier task of picking, it requires a well defined sequence of certain visual perception tasks to be executed. This primarily includes object detection, instance detection and segmentation, instance selection, and instance tracking. While performing these tasks in constrained environments such as uniform background, the perception algorithms turns out be relatively simpler. However, in the scenarios such as outdoors, several uncontrolled variables jumps in, for example, ambient light, dynamic objects, highly complex and dynamic visual scenes, multiple confusing candidates. All such external variables increase the difficulty level to several degrees. Apart from that, real-time inference of perception modules is also desired, however, limited onboard computational and electric power further convolutes the system design.
\subsection{Localization and Navigation}
UAV localization and navigation play an important role in exploring the workspace autonomously and execute waypoint operations. In the GPS based systems, location information can be obtained easily and optionally fused with inertial-measurements-units (IMUs) for high frequency state-estimation. However, in GPS denied situations, the localization and navigation no longer remain an easy task from algorithmic point of view. In such scenarios, visual-SLAM or visual odometery based algorithms are generally employed. These algorithms perform feature matching in images, $3$D point clouds and carryout several optimization procedures which in turn require significant amount of computational power, thus transforming the overall problem into a devil.
\par
Due to the above mentioned complexities, limitations and constraints, performing the task of object manipulation using UAVs is not easy. Hence, in this work, we pave a way to solve the problem of UAV based manipulation in the presence of several above discussed challenges. In this paper, we primarily focus on real-time accurate visual perception, localization and navigation of using visual-SLAM for $6$-DoF state-estimation in GPS-denied environments. The state-estimation is used for several other high level tasks such as autonomous take-off, landing, planning, navigation and control. Later, these modules are utilized to perform an object pick, estimation of successful grasp, transport and place operation which is primarily governed by our deep learning based visual perception pipeline. Highlights of the paper are as follows:
\begin{enumerate}
   \item Identification and in detailed discussion of the challenges associated with UAV based object manipulation.
    \item A novel computationally efficient, deep learning based unified multi-task visual perception system for instance level detection, segmentation, part segmentation, and tracking.
    \item A novel visual learning based grasp state feedback.
    \item A remote computing approach for UAVs.
    \item An electromagnet based gripper design for UAVs.
    \item Developing high precision $6$-DoF state estimation on top of ORB-SLAM$2$ \cite{orb2} visual-SLAM.
\end{enumerate}
\par
A side contributory goal of this paper is to introduce and benefit the community by providing fine details on low level complexities associated with UAV based manipulation and potential solutions to the problem.
% It would likely enable the worldwide researches to get a descent understanding of UAV based manipulation and an entry point for them to come up with much more advanced solutions.
In the next section, we first discuss the available literature. In Sec. \ref{sec_sysdesign}, we discuss the system  design. In Sec. \ref{sec_perception}, we discuss the unified visual perception system. In Sec. \ref{sec_integration}, system integration is discussed. The Sec. \ref{sec_exp} provides the experimental study and Finally, the Sec. \ref{sec_conc} provides the conclusions about the paper.

\section{Related Work}
\label{sec_rel}

Due to diverse literature on robotic vision and very limited space, we are bound to provide very short discussion on relevant work. We introduce only a few popular learning based approaches for the instance detection, segmentation, tracking and visual-SLAM. AlexNet \cite{alexnet}, VGG \cite{vgg}, ResNet \cite{resnet} are very first Convlutional Neural Networks (CNN). The object detectors Faster-RCNN \cite{fasterrcnn}, Fast-RCNN \cite{fastrcnn}, RCNN \cite{rcnn} are developed on top of them. FCN \cite{fcn}, PSPNet \cite{pspnet}, RefineNet \cite{refinenet} are approaches developed for segmentation tasks. Mask-RCNN \cite{maskrcnn} combines \cite{fasterrcnn} and FPN \cite{fpn} to improve object detction accuracies and also proposes an ROI align approach to facilicate instance segmentation. %Panoptic segmentation is very recent apporach for fine-grained semantic segmentation.
\par
ORB-SLAM \cite{orb} and ORB-SLAM$2$ \cite{orb2} are popular approaches for monocular and stereo based visual-SLAM. UnDeepVO \cite{undeepvo} is a very recent approach to visual odometery using unsupervised deep learning. DeepSORT \cite{deepsort} is a deep learning based multi object tracker inspired by kalman filter.
\par All of the above algorithms are being used in robotics worldwise and many recent works revolves around them. Nonetheless, the issue of deploying these algorithms on computationally limited platforms altogether is still a challenge.

\section{System Design}
\label{sec_sysdesign}
\subsection{UAV Platform}
According to our experimental observation, an MAV must have a payload capacity atleast double of the maximum payload to be lifted. It is required in order to avoid complexities involved with dynamic payload attachment (Sec. \ref{subsec_dynpayload}). Keeping this observation in mind, we use DJI Matrice-$600$ Pro hex rotor (Fig. \ref{fig_m600}) which has a payload capacity of $6$ Kgs, flight time of $16$ and $32$ minutes with and without payload respectively. %The UAV comes with triple redundant GPS for better localization. However, we discard GPS data for our purpose.
 The UAV flight controller can be accessed through a UART bus via DJI onboard-SDK APIs. The DJI UAVs are quite popular worldwide for manual flying, however, turning them into autonomous platforms is not straight forward. It is because, being an industrial drone, it is optimized for manual control. Moreover, the SDK APIs lack proper documentation and other important details of low level control, the rate at which the controller can listen commands remain hidden. Availability of these details are quite crucial for robust autonomous control of the UAVs, since a delay of merely $20$ms in the control commands can degrade the performance severely. Despite the challenges, their low cost and decent payload capacity make them attractive choice for research. It encourages us to examine the control responses, delays associated in the communication system of the UAV in order to adapt it for autonomous operations.
\subsection{Gripper Design}
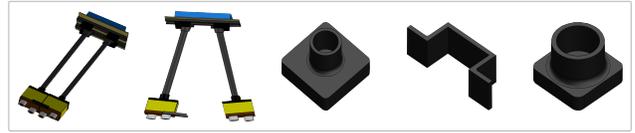
\begin{figure}[t]
\centering
\begin{tikzpicture}
\node () [draw=none,scale = 0.74]{
\tikz{
\node (temp) [draw=none,yshift=0ex]{};
\node (bbox) [draw=white!70!black,below of=temp,rounded corners=0.25mm,yshift=-0.99ex]{
\tikz
{
\node (p1) [rounded corners=1mm,xshift=0ex,scale = 0.5]{\includegraphics[width=25ex,height=25ex]{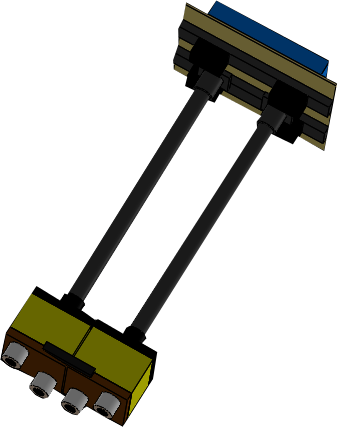}};
\node (p3) [right of=p1,xshift=8ex,rounded corners=1mm,scale = 0.5]{\includegraphics[width=25ex,height=25ex]{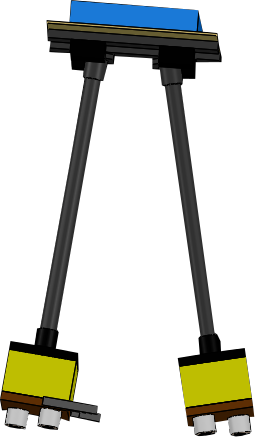}};
\node (p4) [right of = p3,rounded corners=1mm,xshift=8ex,yshift=0ex,scale = 0.4]{\includegraphics[width=25ex,height=25ex]{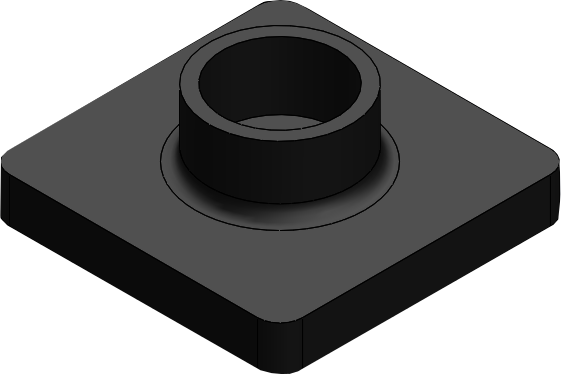}};
\node (p5) [right of=p4,xshift=8ex,yshift=0ex,rounded corners=1mm,scale = 0.4]{\includegraphics[width=25ex,height=25ex]{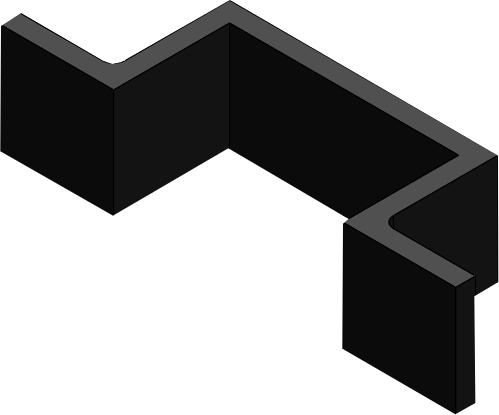}};
\node (p6) [right of=p5,xshift=8ex,yshift=0ex,rounded corners=1mm,scale = 0.4]{\includegraphics[width=25ex,height=25ex]{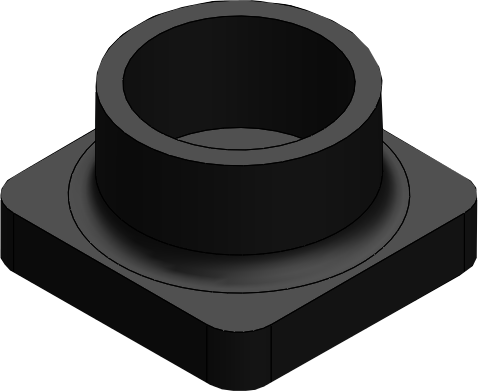}};
}};
}};
\end{tikzpicture}
\caption{The CAD design of our gripper and $3$D printed parts}
\label{fig_gripper}
\end{figure}

We develop a retractable servo actuated gripper (Fig. \ref{fig_gripper}) enabled with electromagnetic grasping. It consists of two carbon fiber tubes, called left and right arms. Robotics grade heavy duty servos are employed to arm and disarm the gripper when in air. Four servos, two on each side are used for high torque and better design stability. Two electromagnets (EMs) on each arm are used for the gripping mechanism. Each arm and respective electromagnets are connected via an assembly which consists of polymer foams sandwiched between carbon fiber plates. The carbon fiber tubes are linked at both ends (servo and Foam assembly) via $3$D printed parts as shown in Fig. \ref{fig_gripper}. While performing a grasping operation, the foam compression/decompression action accounts for the drifts in UAV's actual and desired height. The control circuit and firmware is developed on PIC$18$F$2550$, a Microchip USB-series microcontroller.
\subsection{Compute Infrastructure and Communication Device}
We equip our platform with DJI-Manifold $2$ mini computer, based on NVIDIA Jetson Tegra TX2. The computer is powered by $6$ core CPUs along with $256$ GPU cores, $8$GB RAM and $32$GB eMMC. A dual band $150$ ($2.4$GHz) + $717$ ($5$GHz) = $867$Mbps onboard WiFi is also available. We use MIMO based ASUS ROG RAPTURE GT-AX-$11000$ Tri-Band wireless router, in order to link base station computers with onboard computer. The router offers very high data transfer rates of upto $1148$Mbps on $2.4$GHz and $4802$Mbps on two $5$GHz bands, aggregating upto $11000$Mbps. The motivation behind using such high speed device is to enable real-time high performance remote computing, system monitoring, remote data logging, and algorithmic debugging. The router also make it feasible to develop multiple UAVs based solutions.
% For example, in order to transfer a point cloud of $600\times600$, it requires $4\times3\times600\times600=4$MBs. For real-time speeds, minimum transfer rate of $25-30$Hz is required which results in 120MB$/$s. The onboard $150$Mbps WiFi can provide only $18$MB$/$s in contrast to $600$MB$/$s speeds of router and $200$MB$/$s of the repeater.

%Instead, we use a MIMO based TP-Link AC$2600$ Dual Band WiFi repeater which can provide speeds upto $800$Mbps and $1700$Mbps at $2.4$GHz and $5$GHz bands repectively. The repeater operates at $100-240$VAC and is not suitable for DC supplies. We disassemble the repeater and examine the inbuilt AC-DC converter. It was observed that the repeater operates at 5V DC. Hence, we replace the internal AC-DC converter with LM2596 based adjustable buck converter, operating at 5V, 3A.
%\par
%Finally, we use MIMO based ASUS ROG RAPTURE GT-AX-11000 Tri-Band wireless router in order to link all the computers. The router can provide very high speeds upto $1148$Mbps on $2.4$GHz and $4802$Mbps on two $5$GHz bands, aggregating upto $11000$Mbps. The motivation behind using such high speed devices is to enable real time high performance computing, system monitoring, remote data logging, and algorithmic debugging. For example, in order to transfer a point cloud of $600\times600$, it requires $4\times3\times600\times600=4$MBs. For real-time speeds, minimum transfer rate of $25-30$Hz is required which results in 120MB$/$s. The onboard $150$Mbps WiFi can provide only $18$MB$/$s in contrast to $600$MB$/$s speeds of router and $200$MB$/$s of the repeater.

%
%
\section{Uinified Multi-Task Visual Perception}
\label{sec_perception}
\begin{figure*}[t]
\centering
\begin{tikzpicture}

\node [scale = 0.99]
{
\tikz{
\node (outer_box) [draw=white!70!black,rounded corners=0.25mm,minimum width=110ex, minimum height = 36ex, xshift = -7ex,yshift=-2ex,]{};
%\draw [very thin, dashed] ($(outer_box.west)+(0ex,1.65ex)$) -- ($(outer_box.east)+(0ex,1.65ex)$);
%\draw [very thin, dashed] ($(outer_box.west)-(0ex,11.35ex)$) -- ($(outer_box.east)-(0ex,11.35ex)$);
%\draw [very thin, dashed] ($(outer_box.north)+(3.0ex,-16.5ex)$) -- ($(outer_box.south)+(3.0ex,6.7ex)$);
%
\node (nwarch) [yshift = 1.0ex,xshift=-7ex]
{\tikz{
\node (d1) [fill=none,minimum width=1ex,minimum height=1ex,xshift=-43ex,yshift=16.7ex]{};
\node (d2) [fill=none,minimum width=1ex,minimum height=1ex,xshift=-43ex,yshift=1ex]{};
\node (d3) [fill=none,minimum width=1ex,minimum height=1ex,xshift=14ex,yshift=1ex]{};
\node (d4) [fill=none,minimum width=1ex,minimum height=1ex,xshift=14ex,yshift=5.2ex]{};
\node (d5) [fill=none,minimum width=1ex,minimum height=1ex,xshift=41ex,yshift=5.2ex]{};
\node (d6) [fill=none,minimum width=1ex,minimum height=1ex,xshift=41ex,yshift=16.7ex]{};
\filldraw[fill=white!30!magenta,opacity=0.1, draw=none,line width=0.2ex] (d1.center) --(d2.center) -- (d3.center) -- (d4.center) -- (d5.center) --(d6.center) -- (d1.center);
\draw[opacity=1, draw=magenta,line width=0.2ex,dashdotted] (d1.center) --(d2.center) -- (d3.center) -- (d4.center) -- (d5.center) --(d6.center) -- (d1.center);
\node (d1) [fill=none,minimum width=1ex,minimum height=1ex,xshift=-43ex,yshift=16.7ex]{};
\node (d2) [fill=none,minimum width=1ex,minimum height=1ex,xshift=-43ex,yshift=1ex]{};
\node (d3) [fill=none,minimum width=1ex,minimum height=1ex,xshift=62ex,yshift=1ex]{};
\node (d4) [fill=none,minimum width=1ex,minimum height=1ex,xshift=62ex,yshift=16.7ex]{};
\draw[opacity=1, draw=black, dashdotted,line width=0.2ex] (d1.center) --(d2.center) -- (d3.center) -- (d4.center) -- (d1.center);
\node (d1) [fill=none,minimum width=1ex,minimum height=1ex,xshift=-43ex,yshift=16.7ex]{};
\node (d2) [fill=none,minimum width=1ex,minimum height=1ex,xshift=-43ex,yshift=-11.6ex]{};
\node (d3) [fill=none,minimum width=1ex,minimum height=1ex,xshift=11ex,yshift=-11.6ex]{};
\node (d4) [fill=none,minimum width=1ex,minimum height=1ex,xshift=11ex,yshift=1.0ex]{};
\node (d5) [fill=none,minimum width=1ex,minimum height=1ex,xshift=-15ex,yshift=1.0ex]{};
\node (d6) [fill=none,minimum width=1ex,minimum height=1ex,xshift=-15ex,yshift=16.7ex]{};
\filldraw[fill=white!30!cyan,opacity=0.1, draw=none,line width=0.2ex] (d1.center) --(d2.center) -- (d3.center) -- (d4.center) -- (d5.center) --(d6.center) -- (d1.center);
\draw[opacity=1, draw=cyan,line width=0.2ex,dashdotted] (d1.center) --(d2.center) -- (d3.center) -- (d4.center) -- (d5.center) --(d6.center) -- (d1.center);
\node (d1) [fill=none,minimum width=1ex,minimum height=1ex,xshift=11ex,yshift=-11.6ex]{};
\node (d2) [fill=none,minimum width=1ex,minimum height=1ex,xshift=11ex,yshift=1.0ex]{};
\node (d3) [fill=none,minimum width=1ex,minimum height=1ex,xshift=62ex,yshift=1ex]{};
\node (d4) [fill=none,minimum width=1ex,minimum height=1ex,xshift=62ex,yshift=-11.6ex]{};
\filldraw[fill=white!30!green,opacity=0.1, draw=none,line width=0.2ex] (d1.center) --(d2.center) -- (d3.center) -- (d4.center) -- (d1.center);
\draw[opacity=1, draw=green,line width=0.2ex,dashdotted] (d1.center) --(d2.center) -- (d3.center) -- (d4.center) -- (d1.center);
\FPeval{\imcnnxshift}{0-29}
\FPeval{\imcnnyshift}{9}
\FPeval{\maskcnnxshift}{0-32}
\FPeval{\maskcnnyshift}{0-6}
%
%% imcnn
%
\node (s1) [draw=cyan,minimum width=1.5ex,rounded corners=0.25mm, minimum height = 10ex, xshift = \imcnnxshift ex, yshift = \imcnnyshift ex] {};
\node (s2) [draw=cyan,right of = s1,rounded corners=0.25mm,minimum width=1.5ex, minimum height = 8ex, xshift = -4ex] {};
\node (s3) [draw=cyan,right of = s2,rounded corners=0.25mm,minimum width=1.5ex, minimum height = 6ex, xshift = -4ex] {};
\node (s4) [draw=cyan,right of = s3,rounded corners=0.25mm,minimum width=1.5ex, minimum height = 4ex, xshift = -4ex] {};
\node (s5) [draw=cyan,right of = s4,rounded corners=0.25mm,minimum width=1.5ex, minimum height = 2ex, xshift = -4ex] {};
\node (c1) [draw=none,above of = s1, minimum width=1.5ex,rounded corners=0.25mm, minimum height = 10ex, xshift = 2 ex, yshift = 0 ex,scale=0.6] {$C_1$};
\node (FPN) [draw=green!70!black,right of = s5,minimum width=2.5ex, minimum height = 2ex, xshift = 3 ex]{\scriptsize FPN};
\draw [->,very thin] (s1) -- (s2);
\draw [->,very thin] (s2) -- (s3);
\draw [->,very thin] (s3) -- (s4);
\draw [->,very thin] (s4) -- (s5);
\draw [->,very thin] (s3.north) |- ($(s3.north) + (9.15ex,0.5ex)$) |- ($(FPN.west) + (0ex, 0.75ex)$);   
\draw [->,very thin] (s4.north) |- ($(s4.north) + (5.85ex,0.5ex)$) |- ($(FPN.west) + (0ex, 0.0ex)$); 
\draw [->,very thin] (s5.east) -- ($(s5.east) + (2.0ex,0.0ex)$) -- ($(s5.east) + (2.0ex,-0.75ex)$) -- ($(FPN.west) + (-3ex, -0.75ex)$) -- ($(FPN.west) + (0ex, -0.75ex)$);
%
%
%
%
%
%% maskcnn
%
\node (ms1) [draw=cyan,minimum width=1.5ex,rounded corners=0.25mm, minimum height = 10ex, xshift = \maskcnnxshift ex, yshift = \maskcnnyshift ex] {};
\node (ms2) [draw=cyan,right of = ms1,rounded corners=0.25mm,minimum width=1.5ex, minimum height = 8ex, xshift = -4ex] {};
\node (ms3) [draw=cyan,right of = ms2,rounded corners=0.25mm,minimum width=1.5ex, minimum height = 6ex, xshift = -4ex] {};
\node (ms4) [draw=cyan,right of = ms3,rounded corners=0.25mm,minimum width=1.5ex, minimum height = 4ex, xshift = -4ex] {};
\node (ms5) [draw=cyan,right of = ms4,rounded corners=0.25mm,minimum width=1.5ex, minimum height = 2ex, xshift = -4ex] {};
\node (mc1) [draw=none,above of = ms1, minimum width=1.5ex,rounded corners=0.25mm, minimum height = 10ex, xshift = 5 ex, yshift = -1.5 ex,scale=0.6] {$C_2$};
\draw [->,very thin] (ms1) -- (ms2);
\draw [->,very thin] (ms2) -- (ms3);
\draw [->,very thin] (ms3) -- (ms4);
\draw [->,very thin] (ms4) -- (ms5);
\node (merge1) [draw=orange,right of = ms5,minimum width=1.5ex,rounded corners=0.25mm, minimum height = 2ex, xshift = -4ex] {};
\node (merge2) [draw=orange,right of = merge1,minimum width=1.5ex,rounded corners=0.25mm, minimum height = 4ex, xshift = -4ex] {};
\node (merge3) [draw=orange,right of = merge2,minimum width=1.5ex,rounded corners=0.25mm, minimum height = 6ex, xshift = -4ex] {};
\draw [->,very thin] (s5.south) |- ($(s5.south) + (0.0ex,-4ex)$) -| (merge1.north);   
\draw [->,very thin] (s4.south) -- ($(s4.south) + (0.0ex,-4ex)$) -| (merge2.north);   
\draw [->,very thin] (s3.south) -- ($(s3.south) + (0.0ex,-4ex)$) -|  (merge3.north);   
\draw [->,very thin] (ms5) -- (merge1);
\draw [->,very thin] ($(ms4.east) + (0.0ex,-1.5ex)$) -- ($(merge2.west) + (0.0ex, -1.5ex)$); \draw [->,very thin] ($(ms3.east) + (0.0ex,-2.5ex)$) -- ($(merge3.west) + (0.0ex, -2.5ex)$); 
\node (mFPN) [draw=green!70!black,right of = merge3,minimum width=2.5ex, minimum height = 2ex, xshift = -0.5 ex]{\scriptsize FPN};
\draw [->,very thin] (merge1.south) |- ($(merge1.south) + (7.36ex,-3.5ex)$) |- ($(mFPN.west) + (0ex, -0.75ex)$);    
\draw [->,very thin] (merge2.south) |- ($(merge2.south) + (4.35ex,-2ex)$) |- ($(mFPN.west) + (0ex, -0.0ex)$);    
\draw [->,very thin] ($(merge3.east)+ (0ex,0.75ex)$) -- ($(mFPN.west)+ (0ex, 0.75ex)$);
\node (targetmask) [draw=red!80!orange,right of = merge3,minimum width=2.5ex, minimum height = 2ex, xshift = 7.5 ex]{\scriptsize Target mask};
\draw [->,very thin] (mFPN) -- (targetmask);
%
%
%%%%%% instance detection
\node (RPN) [draw=magenta!70!white,rounded corners=0.25mm, right of = FPN]{\scriptsize RPN};
 \draw [->,very thin] (FPN.east) -- (RPN.west);
 \draw [->,very thin] ($(FPN.east) - (0ex, 0.75ex)$) -- ($(RPN.west) - (0ex,0.75ex)$);
 \draw [->,very thin] ($(FPN.east) + (0ex, 0.75ex)$) -- ($(RPN.west) + (0ex,0.75ex)$);
\node (roipooling) [draw=black!40!white,rounded corners=0.25mm, right of = RPN, xshift=4ex,yshift=4ex, minimum width=13ex, minimum height = 1.5ex]{\scriptsize ROI Pooling $7\times7$};
\node (roialign) [draw=black!40!white,rounded corners=0.25mm, right of = RPN, xshift=4ex,yshift=-4ex, minimum width=13ex, minimum height = 1.5ex]{\scriptsize ROI Align $14\times14$};
\draw [->,very thin] (RPN.north) |- (roipooling.west);
\draw [->,very thin] (RPN.south) |- (roialign.west);
\node (ic1) [draw=cyan,minimum width=1.5ex,rounded corners=0.25mm,right of = roipooling, minimum height = 2ex, xshift = 3ex, yshift = 2ex] {};
\node (ic2) [draw=cyan,right of = ic1,rounded corners=0.25mm,minimum width=1.5ex, minimum height = 2ex, xshift = -4ex] {};
\node (ib1) [draw=cyan,minimum width=1.5ex,rounded corners=0.25mm,right of = roipooling, minimum height = 2ex, xshift = 3ex, yshift = -2ex] {};
\node (ib2) [draw=cyan,right of = ib1,rounded corners=0.25mm,minimum width=1.5ex, minimum height = 2ex, xshift = -4ex] {};
\node (is1) [draw=cyan,minimum width=1.5ex,rounded corners=0.25mm,right of = roialign, minimum height = 2ex, xshift = 3ex, yshift = 2ex] {};
\node (is2) [draw=cyan,right of = is1,rounded corners=0.25mm,minimum width=1.5ex, minimum height = 2ex, xshift = -4ex] {};
\node (ips1) [draw=cyan,minimum width=1.5ex,rounded corners=0.25mm,right of = roialign, minimum height = 2ex, xshift = 3ex, yshift=-2ex] {};
\node (ips2) [draw=cyan,right of = ips1,rounded corners=0.25mm,minimum width=1.5ex, minimum height = 2ex, xshift = -4ex] {};
\draw [->,very thin] (roipooling.north) |- (ic1.west);
\draw [->,very thin] (roipooling.south) |- (ib1.west);
\draw [->,very thin] (ic1) -- (ic2);
\draw [->,very thin] (ib1) -- (ib2);
\draw [->,very thin] (roialign.north) |- (is1.west);
\draw [->,very thin] (roialign.south) |- (ips1.west);
\draw [->,very thin] (is1) -- (is2);
\draw [->,very thin] (ips1) -- (ips2);
\node (classification) [draw=magenta!70!white,rounded corners=0.25mm, right of = ic2,xshift = 7.6ex, yshift = 0ex, minimum width=17ex, minimum height = 1.5ex]{\scriptsize instance classification};
\node (bboxregression) [draw=magenta!70!white,rounded corners=0.25mm, right of = ib2,xshift = 7.6ex, yshift = -0ex, minimum width=17ex, minimum height = 1.5ex]{\scriptsize instance box regression};
\node (instanceseg) [draw=magenta!70!white,rounded corners=0.25mm, right of = is2,xshift = 7.6ex, yshift = -0ex, minimum width=17ex, minimum height = 1.5ex]{\scriptsize instance segmentation};
\draw [->,very thin] (ic2) -- (classification);
\draw [->,very thin] (ib2) -- (bboxregression);
\draw [->,very thin] (is2) -- (instanceseg);
\node (classes) [draw=white!80!black,rounded corners=0.25mm, right of = classification,xshift = 7ex, yshift = 0ex]{
\tikz{
\node [fill=red,xshift = 0ex, sharp corners, xshift = 0ex, yshift =0ex, minimum width=0.5ex, minimum height = 0.5ex,scale = 0.5]{};
\node [fill=green,xshift = 0ex,sharp corners, xshift = 1ex, yshift =0ex, minimum width=0.5ex, minimum height = 0.5ex,scale = 0.5]{};
\node [fill=blue,xshift = 0ex,sharp corners,  xshift = 2ex,yshift =0ex, minimum width=0.5ex, minimum height = 0.5ex,scale = 0.5]{};
\node [fill=orange,xshift = 0ex,sharp corners, xshift = 3ex, yshift =0ex, minimum width=0.5ex, minimum height = 0.5ex,scale = 0.5]{};
}};
\draw [->,very thin] (classification) -- (classes);
\node (boxes) [draw=white!80!black,rounded corners=0.25mm, right of = bboxregression,xshift = 7ex, yshift = 0ex]{
\tikz{
\node [draw=red,xshift = 0ex, sharp corners, xshift = 0ex, yshift =0ex, minimum width=0.5ex, minimum height = 0.5ex,scale = 0.5]{};
\node [draw=green,xshift = 0ex,sharp corners, xshift = 1ex, yshift =0ex, minimum width=0.5ex, minimum height = 0.5ex,scale = 0.5]{};
\node [draw=blue,xshift = 0ex,sharp corners,  xshift = 2ex,yshift =0ex, minimum width=0.5ex, minimum height = 0.5ex,scale = 0.5]{};
\node [draw=orange,xshift = 0ex,sharp corners, xshift = 3ex, yshift =0ex, minimum width=0.5ex, minimum height = 0.5ex,scale = 0.5]{};
}};
\node (masks) [draw=white!80!black,rounded corners=0.25mm, right of = instanceseg,xshift = 7ex, yshift = 0ex]{
\tikz{
\node [draw=red,fill=white!20!black,xshift = 0ex, sharp corners, xshift = 0ex, yshift =0ex, minimum width=0.5ex, minimum height = 0.5ex,scale = 0.5]{};
\node [draw=green,fill=white!20!black,xshift = 0ex,sharp corners, xshift = 1ex, yshift =0ex, minimum width=0.5ex, minimum height = 0.5ex,scale = 0.5]{};
\node [draw=blue,fill=white!20!black,xshift = 0ex,sharp corners,  xshift = 2ex,yshift =0ex, minimum width=0.5ex, minimum height = 0.5ex,scale = 0.5]{};
\node [draw=orange,fill=white!20!black,xshift = 0ex,sharp corners, xshift = 3ex, yshift =0ex, minimum width=0.5ex, minimum height = 0.5ex,scale = 0.5]{};
\node [circle,fill=white,xshift = 0ex, sharp corners, xshift = 0ex, yshift =0ex, minimum width=0.5ex, minimum height = 0.5ex,scale = 0.1]{};
\node [circle,fill=white,xshift = 0ex,sharp corners, xshift = 1ex, yshift =0ex, minimum width=0.5ex, minimum height = 0.5ex,scale = 0.1]{};
\node [circle,fill=white,xshift = 0ex,sharp corners,  xshift = 2ex,yshift =0ex, minimum width=0.5ex, minimum height = 0.5ex,scale = 0.1]{};
\node [circle,fill=white,xshift = 0ex,sharp corners, xshift = 3ex, yshift =0ex, minimum width=0.5ex, minimum height = 0.5ex,scale = 0.1]{};
}};
\draw [->,very thin] (classification) -- (classes);
\draw [->,very thin] (bboxregression) -- (boxes);
\draw [->,very thin] (instanceseg) -- (masks);
\node (mixer) [draw = black, right of = ips2,rounded corners=0.25mm, circle, minimum height = 1ex, minimum width=1ex,xshift = -3ex, scale=0.7 ]{};
\node (mixer_dot) [fill = black,right of = ips2,rounded corners=0.25mm, circle, minimum height = 1ex, minimum width=1ex, xshift = -3ex, scale=0.2]{};
\draw [->,very thin] (ips2) -- (mixer);
 \draw [->,very thin] (masks.south) -- ($(masks.south) - (0ex, 1ex)$) -- ($(mixer.north) + (0ex,1ex)$) -- (mixer.north);
\node (partclassification) [draw=magenta!70!white,rounded corners=0.25mm, right of = mixer,xshift = 4.3ex, yshift = 0ex, minimum width=17ex, minimum height = 1.5ex]{\scriptsize part segmentation};
 \draw [->,very thin] (mixer) -- (partclassification);
\node (partmasks) [draw=white!80!black,rounded corners=0.25mm, right of = partclassification,xshift = 7ex, yshift = 0ex]{
\tikz{
\node [draw=red,fill=yellow,xshift = 0ex, sharp corners, xshift = 0ex, yshift =0ex, minimum width=0.5ex, minimum height = 0.5ex,scale = 0.5]{};
\node [draw=green,fill=yellow,xshift = 0ex,sharp corners, xshift = 1ex, yshift =0ex, minimum width=0.5ex, minimum height = 0.5ex,scale = 0.5]{};
\node [draw=blue,fill=yellow,xshift = 0ex,sharp corners,  xshift = 2ex,yshift =0ex, minimum width=0.5ex, minimum height = 0.5ex,scale = 0.5]{};
\node [draw=orange,fill=yellow,xshift = 0ex,sharp corners, xshift = 3ex, yshift =0ex, minimum width=0.5ex, minimum height = 0.5ex,scale = 0.5]{};
}};
\draw [->,very thin] (partclassification) -- (partmasks);
\node (detectimg) [draw=none,right of = partmasks, xshift = 0ex, sharp corners, xshift = 6ex, yshift =5.5ex, minimum width=0.5ex, minimum height = 0.5ex,scale = 0.3]{\includegraphics[width=22ex,height=22ex]{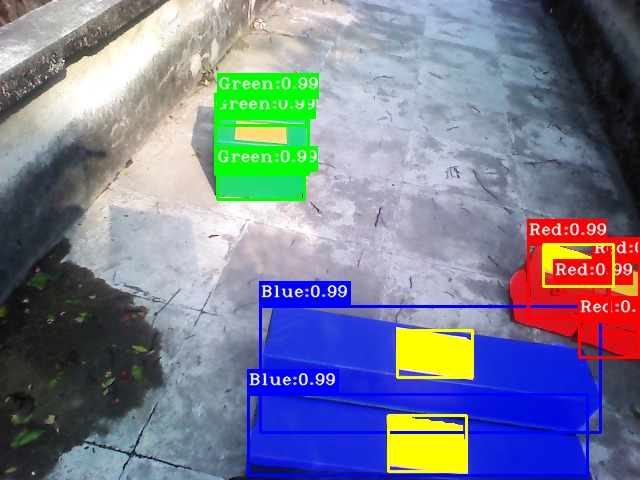}};
\draw [dashed,very thin] (classes) -- (detectimg);
\draw [dashed,very thin] (boxes) -- (detectimg);
\draw [dashed,very thin] (masks) -- (detectimg);
\draw [dashed,very thin] (partmasks) -- (detectimg);
\node (mergeimg) [draw=orange,left of = s1,minimum width=1.5ex,rounded corners=0.25mm, minimum height = 10ex, xshift = 4ex] {};
\draw [->,very thin] (mergeimg) -- (s1);
\node (rgbt1) [draw=none,left of = mergeimg, xshift = 0ex, ,rounded corners=0.25mm, xshift = -1.5ex, yshift =2ex, minimum width=0.5ex, minimum height = 0.5ex,scale = 0.3]{\includegraphics[width=18ex,height=18ex]{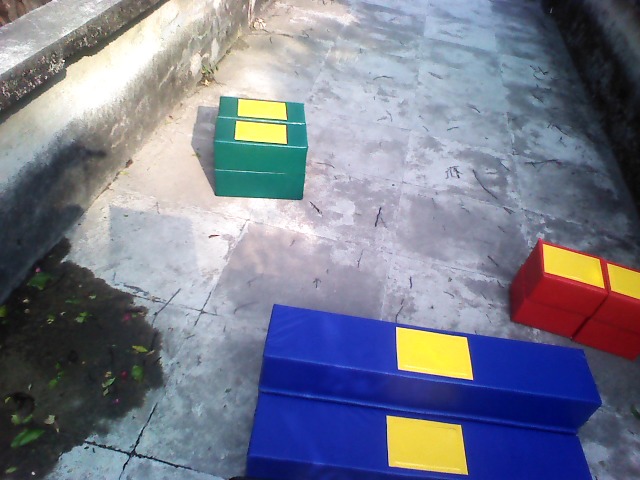}};
\node (rgb) [draw=none,left of = mergeimg, xshift = 0ex, ,rounded corners=0.25mm, xshift = 0ex, yshift =0ex, minimum width=0.5ex, minimum height = 0.5ex,scale = 0.3]{\includegraphics[width=18ex,height=18ex]{nw_arch_img.jpg}};
\draw [->,very thin] (rgb) -- (mergeimg);
\node (rgb_time) [draw=none, below of = rgb, xshift = 0ex, ,rounded corners=0.25mm, xshift = -4ex, yshift =4.3ex, minimum width=0.5ex, minimum height = 0.5ex,scale = 0.8]{$t$};
\node (rgb_time) [draw=none, below of = rgb, xshift = 0ex, ,rounded corners=0.25mm, xshift =-1ex, yshift =2.0ex, minimum width=0.5ex, minimum height = 0.5ex,scale = 0.8]{$t-1$};
\node (mask) [draw=none,left of = ms1,xshift = 0ex, ,rounded corners=0.25mm, xshift = 0ex, yshift =0ex, minimum width=0.5ex, minimum height = 0.5ex,scale = 0.3]{\includegraphics[width=18ex,height=18ex]{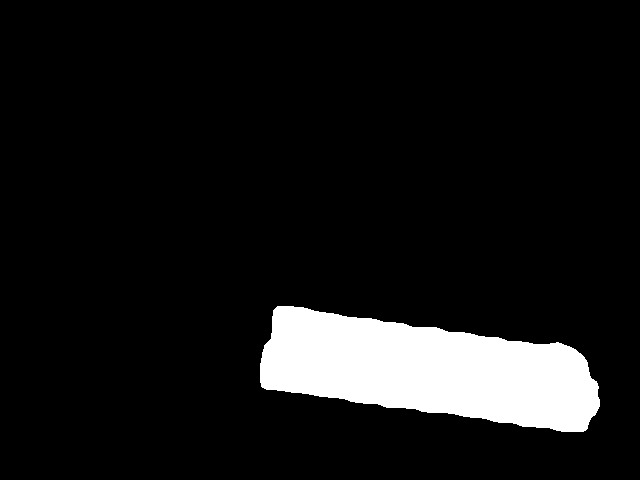}};
\draw [->,very thin] (mask) -- (ms1);
\node (mask_time) [draw=none,below of = mask, xshift = 0ex, ,rounded corners=0.25mm, xshift = 0ex, yshift =2.5ex, minimum width=0.5ex, minimum height = 0.5ex,scale = 0.8]{$t-1$};
\node (maskop) [draw=none,right of = targetmask,xshift = 0ex, ,rounded corners=0.25mm, xshift = 3ex, yshift =0ex, minimum width=0.5ex, minimum height = 0.5ex,scale = 0.3]{\includegraphics[width=18ex,height=18ex]{nw_arch_img_mask.jpg}};
\draw [->,very thin] (targetmask) -- (maskop);
\node (maskop_time) [draw=none,below of = maskop, xshift = 0ex, ,rounded corners=0.25mm, xshift = 0ex, yshift =2.5ex, minimum width=0.5ex, minimum height = 0.5ex,scale = 0.8]{$t$};
}};
\node (graspstate) [draw=none, rounded corners=.25mm, xshift=21ex,yshift = -7.0ex,minimum width=49ex]
{\tikz{
\FPeval{\imcnnxshift}{0-25}
\FPeval{\imcnnyshift}{7}
\FPeval{\maskcnnxshift}{0-25}
\FPeval{\maskcnnyshift}{0-9}
%
%% imcnn
%
\node (s1) [draw=cyan,minimum width=1.5ex,rounded corners=0.25mm, minimum height = 10ex, xshift = \imcnnxshift ex, yshift = \imcnnyshift ex] {};
\node (s2) [draw=cyan,right of = s1,rounded corners=0.25mm,minimum width=1.5ex, minimum height = 8ex, xshift = -4ex] {};
\node (s3) [draw=cyan,right of = s2,rounded corners=0.25mm,minimum width=1.5ex, minimum height = 6ex, xshift = -4ex] {};
\node (s4) [draw=cyan,right of = s3,rounded corners=0.25mm,minimum width=1.5ex, minimum height = 4ex, xshift = -4ex] {};
\node (s5) [draw=cyan,right of = s4,rounded corners=0.25mm,minimum width=1.5ex, minimum height = 2ex, xshift = -4ex] {};
\node (c1) [draw=none,above of = s5,minimum width=1.5ex,rounded corners=0.25mm, minimum height = 10ex, xshift = 0 ex, yshift = -3ex,scale=0.6] {$C_3$};
\draw [->,very thin] (s1) -- (s2);
\draw [->,very thin] (s2) -- (s3);
\draw [->,very thin] (s3) -- (s4);
\draw [->,very thin] (s4) -- (s5);
\node (foam1) [draw=cyan,right of = s5,rounded corners=0.25mm,minimum width=1.5ex, minimum height = 2ex, xshift = -2ex,yshift=2ex] {};
\node (foam2) [draw=cyan,right of = foam1,rounded corners=0.25mm,minimum width=1.5ex, minimum height = 2ex, xshift = -4ex] {};
\node (obj1) [draw=cyan,right of = s5,rounded corners=0.25mm,minimum width=1.5ex, minimum height = 2ex, xshift = -2ex,yshift=-2ex] {};
\node (obj2) [draw=cyan,right of = obj1,rounded corners=0.25mm,minimum width=1.5ex, minimum height = 2ex, xshift = -4ex] {};
\draw [->,very thin] (s5.north) |- (foam1.west);   
\draw [->,very thin] (s5.south) |- (obj1.west);   
\draw [->,very thin] (foam1) -- (foam2);   
\draw [->,very thin] (obj1) -- (obj2);   
\node (foamstateclassification) [draw=magenta!70!white,rounded corners=0.25mm, right of = foam2,xshift = 6.75ex, yshift = 0ex, minimum width=23ex, minimum height = 1ex]{\scriptsize Foam compressed/uncompressed};
\node (objectstateclassification) [draw=magenta!70!white,rounded corners=0.25mm, right of = obj2,xshift = 6.75ex, yshift = -0ex, minimum width=23ex, minimum height = 1ex]{\scriptsize Object Attached/Not Attached};
\draw [->,very thin] (foam2) -- (foamstateclassification);
\draw [->,very thin] (obj2) -- (objectstateclassification);
\node (img) [draw=none,left of=s1,minimum width=1ex,rounded corners=0.25mm, minimum height = 1ex,xshift=2ex,yshift=0ex] {\includegraphics[width=4ex,height=4ex]{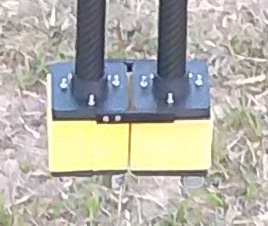}};
\draw [->,very thin] (img) -- (s1);
\draw [->,very thin] (obj2) -- (objectstateclassification);
}};
%
%
%
%
%
% LEGENDS
\node (legends) [draw=none, rounded corners=0.25mm, minimum width=110ex,minimum height=4ex, xshift = -7 ex, yshift = -16.59ex]{
\tikz{
\node (conv) [draw=cyan,minimum width=3.5ex, minimum height =1.5ex,xshift=-100ex]{};
\node (convtext) [draw=white,right of = conv,minimum width=1ex, minimum height =1ex, xshift = 0.8ex,scale=0.8]{\scriptsize Conv-BN-ReLU};
\node (concat) [draw=orange,,rounded corners=0.25mm,minimum width=3.5ex, minimum height =1.5ex,xshift =-85.5ex, yshift = 0ex]{};
\node (concattext) [draw=white,,rounded corners=0.25mm,right of = concat,minimum width=1ex, minimum height =1ex, xshift = 0.5ex,scale=0.8]{\scriptsize Concatenation};
\node (fpn) [draw=green!70!black,,rounded corners=0.25mm,minimum width=3ex, minimum height =1ex,xshift = -71.5ex, yshift = -0.0ex]{\scriptsize FPN};
\node (fpntext) [draw=white,,rounded corners=0.25mm,right of = fpn,minimum width=1ex, minimum height =1ex, xshift = 4.7ex,scale=0.8]{\scriptsize Feature Pyramid Network};
\node (rpn) [draw=magenta!70!white,,rounded corners=0.25mm,minimum width=3ex, minimum height =1ex, xshift= -49.5ex, yshift = 0ex]{\scriptsize RPN};
\node (rpntext) [draw=white,right of = rpn,,rounded corners=0.25mm,minimum width=1ex, minimum height =1ex, xshift = 4.6ex,scale=0.8]{\scriptsize Region Proposal Network};
\node (mixer) [draw = black, circle, minimum height = 1ex, minimum width=1ex,xshift=-29ex,yshift = 0ex, scale=0.7 ]{};
\node (mixer_dot) [fill = black, circle, minimum height = 1ex, minimum width=1ex, xshift = -29ex,yshift=-0.0ex, scale=0.2]{};
\node (mixertext) [draw=white,right of = mixer,minimum width=1ex, minimum height =1ex, xshift = -0.2ex,align=left,scale=0.8]{\scriptsize Eltwise Product};
\node (gstatetxt) [draw=green,rounded corners=0.25mm,minimum width=1ex, minimum height =1ex, xshift = -10ex,dashdotted,line width=0.2ex,scale=0.8]{\scriptsize Grasp State Classifier};
\node (classes) [draw=white!80!black,rounded corners=0.25mm, xshift = -100ex, yshift = -2.8ex,minimum width=1ex,minimum height=1ex]{
\tikz{
\node [fill=red,xshift = 0ex, sharp corners, xshift = 0ex, yshift =0ex, minimum width=0.5ex, minimum height = 0.5ex,scale = 0.5]{};
\node [fill=green,xshift = 0ex,sharp corners, xshift = 1ex, yshift =0ex, minimum width=0.5ex, minimum height = 0.5ex,scale = 0.5]{};
\node [fill=blue,xshift = 0ex,sharp corners,  xshift = 2ex,yshift =0ex, minimum width=0.5ex, minimum height = 0.5ex,scale = 0.5]{};
\node [fill=orange,xshift = 0ex,sharp corners, xshift = 3ex, yshift =0ex, minimum width=0.5ex, minimum height = 0.5ex,scale = 0.5]{};
}};
\node (classtext) [draw=white,right of = classes,minimum width=1ex, minimum height =1ex, xshift = 0.5ex,scale=0.8]{\scriptsize Brick Class};
\node (boxes) [draw=white!80!black,rounded corners=0.25mm,  xshift = -86ex, yshift = -2.8ex,minimum width=1ex,minimum height=1ex]{
\tikz{
\node [draw=red,xshift = 0ex, sharp corners, xshift = 0ex, yshift =0ex, minimum width=0.5ex, minimum height = 0.5ex,scale = 0.5]{};
\node [draw=green,xshift = 0ex,sharp corners, xshift = 1ex, yshift =0ex, minimum width=0.5ex, minimum height = 0.5ex,scale = 0.5]{};
\node [draw=blue,xshift = 0ex,sharp corners,  xshift = 2ex,yshift =0ex, minimum width=0.5ex, minimum height = 0.5ex,scale = 0.5]{};
\node [draw=orange,xshift = 0ex,sharp corners, xshift = 3ex, yshift =0ex, minimum width=0.5ex, minimum height = 0.5ex,scale = 0.5]{};
}};
\node (boxtext) [draw=white,right of = boxes,minimum width=1ex, minimum height =1ex, xshift = 3.2ex,scale=0.8]{\scriptsize Brick Bounding Box};
\node (masks) [draw=white!80!black,rounded corners=0.25mm,  xshift = -66.5ex, yshift = -2.8ex,minimum width=1ex,minimum height=1ex]{
\tikz{
\node [draw=red,fill=white!20!black,xshift = 0ex, sharp corners, xshift = 0ex, yshift =0ex, minimum width=0.5ex, minimum height = 0.5ex,scale = 0.5]{};
\node [draw=green,fill=white!20!black,xshift = 0ex,sharp corners, xshift = 1ex, yshift =0ex, minimum width=0.5ex, minimum height = 0.5ex,scale = 0.5]{};
\node [draw=blue,fill=white!20!black,xshift = 0ex,sharp corners,  xshift = 2ex,yshift =0ex, minimum width=0.5ex, minimum height = 0.5ex,scale = 0.5]{};
\node [draw=orange,fill=white!20!black,xshift = 0ex,sharp corners, xshift = 3ex, yshift =0ex, minimum width=0.5ex, minimum height = 0.5ex,scale = 0.5]{};
\node [circle,fill=white,xshift = 0ex, sharp corners, xshift = 0ex, yshift =0ex, minimum width=0.5ex, minimum height = 0.5ex,scale = 0.1]{};
\node [circle,fill=white,xshift = 0ex,sharp corners, xshift = 1ex, yshift =0ex, minimum width=0.5ex, minimum height = 0.5ex,scale = 0.1]{};
\node [circle,fill=white,xshift = 0ex,sharp corners,  xshift = 2ex,yshift =0ex, minimum width=0.5ex, minimum height = 0.5ex,scale = 0.1]{};
\node [circle,fill=white,xshift = 0ex,sharp corners, xshift = 3ex, yshift =0ex, minimum width=0.5ex, minimum height = 0.5ex,scale = 0.1]{};
}};
\node (masktext) [draw=white,right of = masks,minimum width=1ex, minimum height =1ex, xshift = 3.2ex,scale=0.8]{\scriptsize Brick Instance Mask};
\node (partmasks) [draw=white!80!black,rounded corners=0.25mm,  xshift = -47ex, yshift = -2.8ex,minimum width=1ex,minimum height=1ex]{
\tikz{
\node [draw=red,fill=yellow,xshift = 0ex, sharp corners, xshift = 0ex, yshift =0ex, minimum width=0.5ex, minimum height = 0.5ex,scale = 0.5]{};
\node [draw=green,fill=yellow,xshift = 0ex,sharp corners, xshift = 1ex, yshift =0ex, minimum width=0.5ex, minimum height = 0.5ex,scale = 0.5]{};
\node [draw=blue,fill=yellow,xshift = 0ex,sharp corners,  xshift = 2ex,yshift =0ex, minimum width=0.5ex, minimum height = 0.5ex,scale = 0.5]{};
\node [draw=orange,fill=yellow,xshift = 0ex,sharp corners, xshift = 3ex, yshift =0ex, minimum width=0.5ex, minimum height = 0.5ex,scale = 0.5]{};
}};
\node (partmasktext) [draw=white,right of = partmasks,minimum width=1ex, minimum height =1ex, xshift = 4.2ex,scale=0.8]{\scriptsize Magnetic Region Mask};
\node (ltrackertxt) [draw=cyan,line width=0.2ex,dashdotted,minimum width=1ex,minimum height=1ex,xshift=-24ex, yshift=-2.8ex,scale=0.8]{\scriptsize Tracker};
\node (lmaskrcnntxt) [draw=magenta,line width=0.2ex,dashdotted,minimum width=1ex,minimum height=1ex,xshift=-12ex, yshift=-2.8ex,scale=0.8]{\scriptsize Mask-RCNN};
\node (ldetectortxt) [draw=black,line width=0.2ex,dashdotted,minimum width=1ex,minimum height=1ex,xshift=0ex, yshift=-2.8ex,scale=0.8]{\scriptsize Detector};
}};
}};
\end{tikzpicture}
\caption{Unified Visual Perception System}
\label{fig_cnn}
\end{figure*}
\par
Deep learning based visual perception systems perform with very high accuracy on tasks such as object detection, instance detection, segmentation, visual tracking. All of these methods exploit the power of CNNs to generate unique high dimensional latent representation of the data. During the past decade, a number of CNN based algorithms have been proposed which have shown increasing improvements in these tasks over time. The CNN models of these algorithms are well tested on standard datasets, however, lacks generalization. These models, often, are also very large in size. Therefore, direct deployment of such algorithms for real-time robotic applications is not a wise choice. 
\par
As an example of our case, four tasks i.e. instance detection, segmentation, part segmentation and object tracking need to be performed in order to execute a high level task. It is quite important to note that despite the availability of several algorithms and their open source implementations individually, none of them combines these primary tasks together.
\par
Therefore, in the realm of limited computational and power resources, we put an effort to combine aforementioned four tasks in a unified CNN powered visual perception framework. Instead of developing perception modules from scratch, we develop the unified multi-task visual perception system on the top of the state-of-art algorithms. It is done in order to achieve the goals of computational and resource efficiency, real-time, robust and fail-safe performance. 
\par
Being a highly complex visual perception pipeline, we try our best to explain the overall perception system pictorially in Fig. \ref{fig_cnn}. Below we have discussed each of the component in detail.
\subsection{Instance Detection and Segmentation}
In the presence of multiple instances of an object, it is necessary to perform instance level detection and segmentation. Mask-RCNN \cite{maskrcnn} is one of the popular choices for this purpose. The pretrained models of Mask-RCNN utilize ResNet-$50$, ResNet-$101$ as CNN backbones for large datasets such as MS-COCO, PASCAL-VOC. Run time performance of this algorithm is limited to $2$-$5$ FPS even on a high-end GPU device having $\sim3500$ GPU cores. Hence, deploying even a baseline version of Mask-RCNN on Jetson TX$2$ category boards appears as a major bottleneck in algorithmic design.
\par
As a solution to adapt Mask-RCNN for our system, we carefully develop an AlexNet style five stage CNN backbone with MobileNet-v1 style depthwise spearable convolution. After several refinements of parameter selection, we come up with a very small and lightweight model which can run in real-time. In the baseline Mask-RCNN, object detection is performed upto stage-$2$ of ResNet. However, due to resource limitation, we limit the object detection upto stage-3 starting from stage-$5$. Rest of the architecture for instance detection, segmentation remains same.
\par
Further, in order to improve the accuracy, we first train the final model on mini ImageNet ILSVRC-$2012$ dataset so that primary layers of the model can learn meaningfull edge and color responses similar to primary visual cortex in biological vision system. We then fine tune the model for our task of instance detection and segmentation. Pretraining on ImageNet improves the accuracy, generalization and suppresses false positives.
\subsection{Part Detection}
The Mask-RCNN is only limited to the task of instance detection and segmentation. However, we have an additional requirement to localize specific part of an object, in our case, the ferromagnetic regions. Therefore, we extend the previously developed CNN infrastructure to accommodate this functionality. In order to achieve that, features from ROI-Align layer are passed through a stack of two convolution layers, similar to instance segmentation branch (Fig. \ref{fig_cnn}). The features corresponding to an instance are then multiplied elementwise with the segmentation mask of the same instance obtained from instance segmentation branch. Mask-RCNN follows the object centric approach for instance segmentation and performs binary classification using binary cross entropy loss. We also follow the same approach to learn binary mask. For more detailed explanation, please refer to \cite{maskrcnn}.
\subsection{Target and Conditional Tracking and Visual Servoing}
Once the instances of desired class are detected and segmented, the instance requiring minimal translation control efforts is selected for grasping operation. A binary mask $M_{t-1}$ corresponding to the selected target is sent to the tracker. The overall tracking process is conditioned on $M_{t-1}$ where the non-zeros pixels of $M_{t-1}$ guides the tracker ``where to look''. For this reason, we term the target mask as support mask. The task of tracker thus can be defined as to predict support mask image $M_t$ at current time step given current $I_t$ and previous $I_{t-1}$ RGB frames and support mask $M_{t-1}$.
\par
We further extend the developed perception system so far to be used as lightweight tracker (Fig. \ref{fig_cnn}). In order to realize the the tracking process, we design a separate AlexNet style CNN ($C_2$) (similar to $C_1$ and) to extract spatial feature embeddings of the support mask image. The backbone $C_2$ has negligible number of weights as compared to the backbone $C_1$ for RGB images. We have designed the architecture such that both the tracker and the detector share a common CNN backbone $C_1$ for high dimensional embedding of RGB images. This is done in order to prevent the computational resources (limited GPU memory) from being exhausted.
\par
Next, feature embedding of both RGB images and the support mask are fused by first performing a concatenation operation on the embeddings of stage-$3$, stage-$4$ and stage-$5$ of both $C_1$ and $C_2$. This step essentially represents the conditioning of the tracking process onto $M_{t-1}$. Later, these fused representations are aggregated by using FPN in order to predict highly detailed support mask for next time step. Despite the very small size of CNN architecture, the FPN module is responsible \cite{deepquick} for highly detailed support mask. Higher resolution embeddings are avoided due to overly large memory requirements.
%\par
%In tracking process, temporal history about the target object such as location, speed, appearance should be available in order to make accurate future predictions. The predictions based on larger history requires longer time, therefore, we reduce the history length to only one previous frame for RGB image as wella single frame for mask image.
\par
To begin with the tracking process, first an instance in an image $I_{t-1}$ is selected and a corresponding binary mask $M_{t-1}$ is obtained. Later, $I_t, I_{t-1}$ and $M_{t-1}$ are fed to their respective CNNs $C_1,C_2$ of the tracker. A binary mask $M_t$ is predicted by the tracker which depicts the location of the target object at time instant $t$. This newly estimated location of the target instance is then sent to the control system which perform real time visual servoing in order to execute a pick operation.
\par
Further, it is important to note that while tracking, the detection modules following the FPN block of the detector (Fig. \ref{fig_cnn}) are halted to save computations. The detection is generally performed on a single Image while tracking process requires two images to be fed to $C_1$. As mentioned previously, the detector and tracker shares a common backbone, therefore $C_1$ must be provided with two images regardless of the task. Therefore, $I_{t-1}$ and $I_t$ are essentially a copy of each other in the detection while these two are different during the tracking process.

\subsection{Learning Based Feedback for Robust Grasping}
\label{subsec_graspstate}
In order to execute a robust grasping operation, continuous feedback of the grasp state i.e. ``gripper contact with object'', ``object attached with gripper'' are required. The feedback device typically depends on the nature of gripper. For example,  finger gripper generally use force sensors to sense the grasp state. In our case, EMs are used which requires dedicated circuitry to obtain the feedbacks such as ``whether the EMs are in contact with any ferromagnetic material''. It poses additional design constraints and system gets extremely complicated.
\par
To simplify the design, we translate the problem of obtaining grasp state feedback into image classification problem. In order to achieve that we take advantage of the fact that the foam assembly of our gripper gets compressed at the moment of contact with the object. In order to realize the approach, We develop a five stage CNN $C_3$ (Fig. \ref{fig_gripper}) where the stage-5 is followed by two classification heads. One of them classifies an image into $G_1$ or $G_2$ whereas another classifies into $G_3$ or $G_4$. 
\par
The four classes $G_1,G_2,G_3,G_4$ represent the states ``Foam Compressed'', ``Foam Uncompressed'', ``Object Attached'' and ``Object Not-Attached'' respectively. The input image is captured from the downward facing Intel realsense D435i mounted on the gripper. Due to the fixed rigid body transformation between gripper and camera, the gripper remains visible at a constant place in the image. Therefore, input to the classifier is a cropped image. The cropping is done such that foam, the electromagnets and the ferromagnetic regions (in case object is attached) remains image centered. The cropped image is resized to $64\times64$ before feeding to the classifier. For the data collection and training details are provided in Sec. \ref{sec_exp}.
\par
At the time of performing grasp operation, the UAV descents at very low speed with the help of visual servoing until the grasp state classifier predicts $G_2$, also referred as touch down. Once the touch down is received, the UAV starts ascending slowly while the grasp state classifier continuously monitors the gripper in this phase. Once the classifier confirms successfull grasp which is denoted by $G_3$, the UAV continues to ascent, transports and places the brick at desired location.
\section{System Integration}
\label{sec_integration}
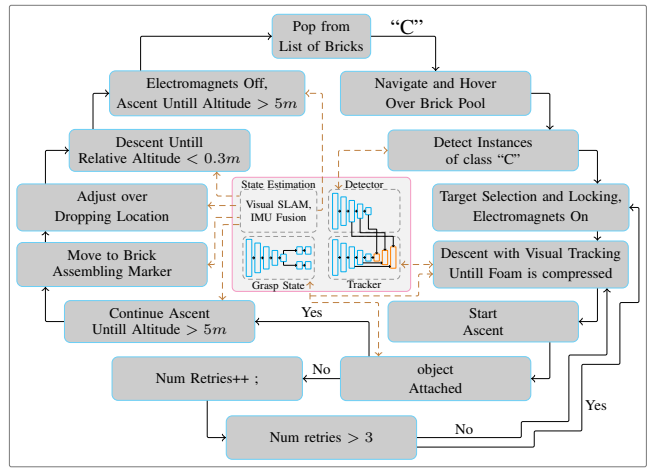
\begin{figure}[t]
\centering
\begin{tikzpicture}
\node [fill=none, draw=white!50!black,rounded corners=0.25mm, scale = 0.8]
{
\tikz{
\node (processing) [scale=0.75, yshift=0ex,xshift=0ex]{
\tikz{
\node (outer_box) [draw=white!50!magenta,fill=white!95!black,rounded corners=1mm,minimum width=25ex, minimum height = 16ex, xshift = 0ex]{};
%\draw [very thin, dashed] (outer_box.west) -- (outer_box.east);
%\draw [very thin, dashed] (outer_box.north) -- (outer_box.south);
%
\node (outer_box) [draw=white!60!black,rounded corners=1mm,minimum width=10ex, minimum height = 6ex, xshift = 6ex,yshift=3.3ex,densely dashed,label={[label distance = -0.7ex]90:\scriptsize Detector}]{};
\node (outer_box) [draw=white!60!black,rounded corners=1mm,minimum width=10ex, minimum height = 6ex, xshift = 6ex,yshift=-3.3ex,densely dashed,label={[label distance = -8.1ex]90:\scriptsize Tracker}]{};
\node (nwarch) [yshift = 1.2ex,xshift=6ex, yshift=-1.2ex, scale = 0.5]
{\tikz{
\FPeval{\imcnnxshift}{0-25}
\FPeval{\imcnnyshift}{7}
\FPeval{\maskcnnxshift}{0-25}
\FPeval{\maskcnnyshift}{0-6}
%
%% imcnn
%
\node (s1) [draw=cyan,minimum width=1.5ex,rounded corners=0.25mm, minimum height = 10ex, xshift = \imcnnxshift ex, yshift = \imcnnyshift ex] {};
\node (s2) [draw=cyan,right of = s1,rounded corners=0.25mm,minimum width=1.5ex, minimum height = 8ex, xshift = -4ex] {};
\node (s3) [draw=cyan,right of = s2,rounded corners=0.25mm,minimum width=1.5ex, minimum height = 6ex, xshift = -4ex] {};
\node (s4) [draw=cyan,right of = s3,rounded corners=0.25mm,minimum width=1.5ex, minimum height = 4ex, xshift = -4ex] {};
\node (s5) [draw=cyan,right of = s4,rounded corners=0.25mm,minimum width=1.5ex, minimum height = 2ex, xshift = -4ex] {};
\draw [->,very thin] (s1) -- (s2);
\draw [->,very thin] (s2) -- (s3);
\draw [->,very thin] (s3) -- (s4);
\draw [->,very thin] (s4) -- (s5);
%
%
%% maskcnn
%
\node (ms1) [draw=cyan,minimum width=1.5ex,rounded corners=0.25mm, minimum height = 10ex, xshift = \maskcnnxshift ex, yshift = \maskcnnyshift ex] {};
\node (ms2) [draw=cyan,right of = ms1,rounded corners=0.25mm,minimum width=1.5ex, minimum height = 8ex, xshift = -4ex] {};
\node (ms3) [draw=cyan,right of = ms2,rounded corners=0.25mm,minimum width=1.5ex, minimum height = 6ex, xshift = -4ex] {};
\node (ms4) [draw=cyan,right of = ms3,rounded corners=0.25mm,minimum width=1.5ex, minimum height = 4ex, xshift = -4ex] {};
\node (ms5) [draw=cyan,right of = ms4,rounded corners=0.25mm,minimum width=1.5ex, minimum height = 2ex, xshift = -4ex] {};
\draw [->,very thin] (ms1) -- (ms2);
\draw [->,very thin] (ms2) -- (ms3);
\draw [->,very thin] (ms3) -- (ms4);
\draw [->,very thin] (ms4) -- (ms5);
\node (merge1) [draw=orange,right of = ms5,minimum width=1.5ex,rounded corners=0.25mm, minimum height = 2ex, xshift = -4ex] {};
\node (merge2) [draw=orange,right of = merge1,minimum width=1.5ex,rounded corners=0.25mm, minimum height = 4ex, xshift = -4ex] {};
\node (merge3) [draw=orange,right of = merge2,minimum width=1.5ex,rounded corners=0.25mm, minimum height = 6ex, xshift = -4ex] {};
\draw [->,very thin] (s5.south) -- ($(s5.south) + (0.0ex,-4ex)$) -- ($(s5.south) + (2.34ex,-4ex)$) -- (merge1.north);   
\draw [->,very thin] (s4.south) -- ($(s4.south) + (0.0ex,-4ex)$) -- ($(s4.south) + (6.96ex,-4ex)$) -- (merge2.north);   
\draw [->,very thin] (s3.south) -- ($(s3.south) + (0.0ex,-4ex)$) -- ($(s3.south) + (11.61ex,-4ex)$) -- (merge3.north);   
\draw [->,very thin] (ms5) -- (merge1);
\draw [->,very thin] ($(ms4.east) + (0.0ex,-1.5ex)$) -- ($(merge2.west) + (0.0ex, -1.5ex)$); \draw [->,very thin] ($(ms3.east) + (0.0ex,-2.5ex)$) -- ($(merge3.west) + (0.0ex, -2.5ex)$); 
}};
\node (uavstate) [draw=white!60!black,rounded corners=1mm,minimum width=10ex, minimum height = 6ex, xshift = -6ex,yshift=3.3ex,densely dashed,label={[label distance = -0.7ex]90:\scriptsize State Estimation},scale = 1]{\shortstack{\scriptsize Visual SLAM, \\ \scriptsize IMU Fusion}};
\node (outer_box) [draw=white!60!black,rounded corners=1mm,minimum width=10ex, minimum height = 6ex, xshift = -6ex,yshift=-3.3ex,densely dashed,label={[label distance = -8.5ex]90:\scriptsize Grasp State}]{};
\node (graspstate) [draw=none, rounded corners=.25mm, xshift=-6ex,yshift = -3.3ex,scale=0.52]
{\tikz{
\FPeval{\imcnnxshift}{0-25}
\FPeval{\imcnnyshift}{7}
\FPeval{\maskcnnxshift}{0-25}
\FPeval{\maskcnnyshift}{0-9}
%
%% imcnn
%
\node (s1) [draw=cyan,minimum width=1.5ex,rounded corners=0.25mm, minimum height = 10ex, xshift = \imcnnxshift ex, yshift = \imcnnyshift ex] {};
\node (s2) [draw=cyan,right of = s1,rounded corners=0.25mm,minimum width=1.5ex, minimum height = 8ex, xshift = -4ex] {};
\node (s3) [draw=cyan,right of = s2,rounded corners=0.25mm,minimum width=1.5ex, minimum height = 6ex, xshift = -4ex] {};
\node (s4) [draw=cyan,right of = s3,rounded corners=0.25mm,minimum width=1.5ex, minimum height = 4ex, xshift = -4ex] {};
\node (s5) [draw=cyan,right of = s4,rounded corners=0.25mm,minimum width=1.5ex, minimum height = 2ex, xshift = -4ex] {};
\draw [->,very thin] (s1) -- (s2);
\draw [->,very thin] (s2) -- (s3);
\draw [->,very thin] (s3) -- (s4);
\draw [->,very thin] (s4) -- (s5);
\node (foam1) [draw=cyan,right of = s5,rounded corners=0.25mm,minimum width=1.5ex, minimum height = 2ex, xshift = -2ex,yshift=2ex] {};
\node (foam2) [draw=cyan,right of = foam1,rounded corners=0.25mm,minimum width=1.5ex, minimum height = 2ex, xshift = -4ex] {};
\node (obj1) [draw=cyan,right of = s5,rounded corners=0.25mm,minimum width=1.5ex, minimum height = 2ex, xshift = -2ex,yshift=-2ex] {};
\node (obj2) [draw=cyan,right of = obj1,rounded corners=0.25mm,minimum width=1.5ex, minimum height = 2ex, xshift = -4ex] {};
\draw [->,very thin] (s5.north) |- (foam1.west);   
\draw [->,very thin] (s5.south) |- (obj1.west);   
\draw [->,very thin] (foam1) -- (foam2);   
\draw [->,very thin] (obj1) -- (obj2);   
}};
}};
\node (pop) [draw=white!50!cyan,xshift=0ex,yshift=20ex,xshift=0ex,yshift=1ex,fill=white!80!black,rectangle,rounded corners=1mm,minimum width=10ex,minimum height=1ex]{\shortstack{\scriptsize Pop from \\ \scriptsize List of Bricks}};
\node (brickpool) [draw=white!50!cyan,xshift=12ex,yshift=15ex,fill=white!80!black,rectangle,rounded corners=1mm,minimum width=20ex,minimum height=3ex,align=center]{\shortstack{\scriptsize Navigate and Hover\\ \scriptsize Over Brick Pool}};
\node (emoffsafealt) [draw=white!50!cyan,xshift=-12ex,yshift=15ex,fill=white!80!black,rectangle,rounded corners=1mm,minimum width=20ex,minimum height=3ex,text justified, align=center]{\shortstack{\scriptsize Electromagnets Off, \\ \scriptsize Ascent Untill Altitude $>5m$}};
\node (detectobject) [draw=white!50!cyan,xshift=17ex,yshift=9ex,fill=white!80!black,rectangle,rounded corners=1mm,minimum width=20ex,minimum height=3ex,align=center]{\shortstack{\scriptsize Detect Instances \\ \scriptsize of class ``C''}};
\node (deliver) [draw=white!50!cyan,xshift=-17ex,yshift=9ex,fill=white!80!black,rectangle,rounded corners=1mm,minimum width=19ex,minimum height=2ex,text justified, align=center]{\shortstack{\scriptsize Descent Untill\\ \scriptsize Relative Altitude $<0.3m$}};
\node (selecttargte) [draw=white!50!cyan,xshift=22ex,yshift=3ex,fill=white!80!black,rectangle,rounded corners=1mm,minimum width=20ex,minimum height=2ex,align=center]{\shortstack{\scriptsize Target Selection and Locking, \\ \scriptsize Electromagnets On}};
\node (adjustoverdrop) [draw=white!50!cyan,xshift=-22ex,yshift=3ex,fill=white!80!black,rectangle,rounded corners=1mm,minimum width=20ex,minimum height=2ex,text justified, align=center]{\shortstack{\scriptsize Adjust over\\ \scriptsize Dropping Location}};
\node (foamstatus) [draw=white!50!cyan,xshift=22ex,yshift=-3ex,fill=white!80!black,rectangle,rounded corners=1mm,minimum width=20ex,minimum height=3ex,text width=19ex,align=center]{\shortstack{\scriptsize Descent with Visual Tracking \\ \scriptsize Untill Foam is compressed}};
\node (movetowall) [draw=white!50!cyan,xshift=-22ex,yshift=-3ex,fill=white!80!black,rectangle,rounded corners=1mm,minimum width=20ex,minimum height=2ex,text justified, align=center]{\shortstack{\scriptsize Move to Brick\\ \scriptsize Assembling Marker}};
\node (ascent) [draw=white!50!cyan,xshift=17ex,yshift=-9ex,fill=white!80!black,rectangle,rounded corners=1mm,minimum width=20ex,minimum height=3ex,align=center]{\shortstack{\scriptsize Start \\ \scriptsize Ascent}};
\node (continueascent) [draw=white!50!cyan,xshift=-17ex,yshift=-9ex,fill=white!80!black,rectangle,rounded corners=1mm,minimum width=20ex,minimum height=2ex,text justified, align=center]{\shortstack{\scriptsize Continue Ascent \\ \scriptsize Untill Altitude $>5m$}};
\node (objectattached) [draw=white!50!cyan,xshift=12ex,yshift=-15ex,fill=white!80!black,rectangle,rounded corners=1mm,minimum width=20ex,minimum height=2ex,text justified, align=center]{\shortstack{\scriptsize object\\ \scriptsize Attached}};
\node (retries) [draw=white!50!cyan,xshift=-12ex,yshift=-15ex,fill=white!80!black,rectangle,rounded corners=1mm,minimum width=20ex,minimum height=4.57ex,text justified, align=center]{\shortstack{\scriptsize Num Retries++ ;}};
\node (retriescheck) [draw=white!50!cyan,xshift=0ex,yshift=-21.2ex,fill=white!80!black,rectangle,rounded corners=1mm,minimum width=20ex,minimum height=4.57ex,text justified, align=center]{\shortstack{\scriptsize Num retries $>3$}};
\draw [->,thin] (pop) -| (brickpool) node [xshift=-3ex,yshift=7ex]{``C''};
\draw [->,thin] (brickpool) -| ($(detectobject.north)+(7ex,0ex)$);
\draw [->,thin] (detectobject) -| ($(selecttargte.north)+(7ex,0ex)$);
\draw [->,thin] ($(selecttargte.south)+(7ex,0ex)$) -- ($(foamstatus.north)+(7ex,0ex)$);
\draw [->,thin] ($(foamstatus.south)+(7ex,0ex)$) |- (ascent);
\draw [->,thin] ($(ascent.south)+(7ex,0ex)$) |- (objectattached);
\draw [->,thin] (objectattached) -- (retries) node [xshift=12ex, yshift=1ex]{\scriptsize No};
\draw [->,thin] ($(objectattached.north)-(7ex,0ex)$) |- (continueascent) node [xshift=16ex, yshift=1ex]{\scriptsize Yes};
\draw [->,thin] (retries) |- (retriescheck);
\draw [->,thin] (retriescheck) -| ($(foamstatus.south)+(-1ex,-14ex)$) -- ($(foamstatus.south)+(4ex,-14ex)$) -- ($(foamstatus.south)+(4ex,-7ex)$) -- ($(foamstatus.south)+(8ex,-7ex)$) -- ($(foamstatus.south)+(8ex,0ex)$) node [xshift=-15ex, yshift=-14.8ex]{\scriptsize No};
\draw [->,thin] ($(retriescheck.east)-(0ex,1ex)$) -| ($(selecttargte.east)+(-10ex,-23.5ex)$) -| ($(selecttargte.east)+(-5ex,-16.5ex)$) -- ($(selecttargte.east)+(-1ex,-16.5ex)$) -- ($(selecttargte.east)+(-1ex,-10ex)$) -- ($(selecttargte.east)+(1ex,-10ex)$) |- ($(selecttargte.east)+(0ex,0ex)$) node [xshift=-3.5ex, yshift=-20.8ex]{\scriptsize Yes};
\draw [->,thin] (continueascent) -| ($(movetowall.south)-(7ex,0ex)$);
\draw [->,thin] ($(movetowall.north)-(7ex,0ex)$) -| ($(adjustoverdrop.south)-(7ex,0ex)$);
\draw [->,thin] ($(adjustoverdrop.north)-(7ex,0ex)$) |- (deliver);
\draw [->,thin] ($(deliver.north)-(7ex,0ex)$) |- (emoffsafealt);
\draw [->,thin] ($(emoffsafealt.north)-(7ex,0ex)$) |- (pop);
\node (detectorlink) [draw=none,fill=none,xshift=1.8ex,yshift=4.15ex]{};
\draw [<->,thin, densely dashed, brown] (detectorlink) |- (detectobject);
\node (trackerlink) [draw=none,fill=none,xshift=7.6ex,yshift=-3ex]{};
\draw [<->,thin, densely dashed, brown] (trackerlink) -- (foamstatus);
\node (graspcnnlink) [draw=none,fill=none,xshift=-1.2ex,yshift=-4ex]{};
\draw [<->,thin, densely dashed, brown] (graspcnnlink) -- ($(graspcnnlink)-(0ex,3ex)$) -| ($(objectattached.north)-(6ex,0ex)$);
\draw [->,thin, densely dashed, brown] ($(graspcnnlink)-(0ex,2.3ex)$) -|  ($(foamstatus.west)-(1.5ex,1ex)$) -- ($(foamstatus.west)-(0ex,1ex)$) ;
\node (stateest1) [draw=none,fill=none,xshift=-1.2ex,yshift=2.7ex]{};
\draw [->,thin, densely dashed, brown] (stateest1) -- ($(stateest1)+(1.3ex,0ex)$) |- (emoffsafealt.east);
\node (stateest2) [draw=none,fill=none,xshift=-5ex,yshift=2.7ex]{};
\draw [<-,thin, densely dashed, brown] ($(deliver.south)+(6ex,0ex)$) |- ($(stateest1)-(7.4ex,-1.5ex)$);
\draw [<-,thin, densely dashed, brown] ($(adjustoverdrop.east) +(0ex,0.20ex)$) -- ($(stateest1)-(7.3ex,-0.50ex)$);
\draw [<-,thin, densely dashed, brown] (movetowall.east) -|($(stateest1)-(10ex,0.75ex)$) -- ($(stateest1)-(7.4ex,0.75ex)$);
\draw [<-,thin, densely dashed, brown] ($(continueascent.north)+(6.6ex,0ex)$) |- ($(stateest1)-(7.4ex,1.5ex)$);
}};
\end{tikzpicture}
\caption{Simplified State Machine of the overall System}
\label{fig_state_machine}
\end{figure}
\subsection{Sensor Fusion and State Estimation}
In order to navigate and perform various tasks accurately in a workspace, the UAV must be aware of its $6$D pose i.e. $\{x,y,z,r,p,y\}$ at every instant of time. To address this issue in GPS-Denied environments, we develop a state-estimation solution which can provide a real-time state-estimation despite limited computational resources onboard the UAV.
\par
The solution is based on very popular real time ORB-SLAM$2$ \cite{orb2} algorithm. ORB-SLAM$2$ can work on stereo as well as RGB-D sensors. We prefer to use stereo based SLAM over RGB-D SLAM because of three reasons. First, RGB-D sensors quite costly as compared to RGB cameras and most of them does not work in outdoor. Second, RGB-D sensors have several points with missing depth information. Third, RGB-D cameras have limited depth range. On the other hand, in stereo based SLAM, the depth of even far points can be computed provided adequate baseline separation between cameras. However, the accuracy of stereo based SLAM comes at a cost which is related to time consumed in keypoint detection, stereo matching, bundle adjustment procedure. The standard implementation of ORB-SLAM2 can process images of resolution $320\times240$ at $25$ FPS on a core $i7$ CPU. When it is deployed onto the Jetson-TX$2$ along with other processes, the speed drops to $5$FPS.
\par
With the limited power budget, it is not possible to place another system alongside the TX$2$ board. Therefore, in order to solve this issue, we take advantage of our high speed networking infrastructure. In order to execute the SLAM algorithm, the captured stereo images are sent to a base station. The SLAM algorithm is executed on the received images to update the $6$D UAV state and $3$D map. As soon as $6$D pose from SLAM is obtained, it is fused with IMU by using Extended Kalman Filter in order to obtain a high frequency state estimate.% It is because the cameras generally operate at 30 FPS while the IMUs can provide data at very high rate (~$100-400$ Hz). Further, In order to guarantee real time image transfer over the network, we use Lighweight-Communication and Marshalling (LCM) library which is generally used in industries for autonomous vechiles.
\subsection{Path Planning and Collision Avoidance}
We use RRTConnect* algorithm for real time planning and flexible collision library (FCL) library for collision avoidance. In order to use them, we modify an open source project MoveIt!, which is was originally developed for robotic manipulators and already has support for RRTConnect* and  FCL. However, MoveIt! has several bugs related to planning of $6$D joints\footnote{``Floating Joint'' in the context of MoveIt!}. For our use, we resolve the issues and configure MoveIt! for UAV. 
\subsection{Control}
As mentioned previously, DJI SDK APIs does not expose its internal controller details. It also doesn't allow the state-estimation sent to its internal controllers. Therefore,  We tune four outer loop PIDs, one for each of vertical velocity, roll and pitch angle and yaw velocity respectively. Based on state-estimation, the tuned PIDs performs position control of the UAV by producing commands for internal hidden controllers,
\par
Fig. \ref{fig_state_machine} shows simplified state machine of our UAV based manipulation system. For the sake of brevity, very obvious steps have been dropped from the state machine. It can be noticed that the state machine can be completely explained by the visual perception, state-estimation, planning and control algorithms discussed previously. 

\section{Experiments}
\label{sec_exp}
\begin{figure*}[t]
\centering
\subfloat[]
{
\begin{tikzpicture}

\node (core) [scale=0.46]{
\tikz{
\node (outer) [scale=1]{
\tikz{
%
%\colorlet{fillcolor}{white!90!black}
\colorlet{fillcolor}{white!98!black}
\foreach \j in {1,2,3,4}
{
\foreach \i in {0,1,...,7}
{
\node (t\j\i) [draw=none,rounded corners, xshift=\i*14ex, yshift=-(\j-1)*13.3ex]{\includegraphics[width=13.5ex,height=12.5ex]{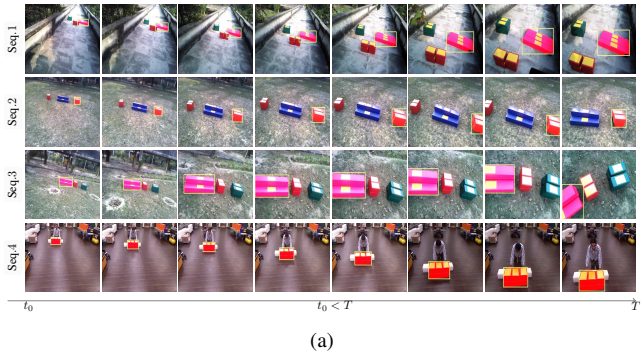}};
}
\node (seq\j) [draw=white,fill=fillcolor,rounded corners=0.6mm,xshift=-9ex, yshift=-(\j-1)*13.3ex, rotate=90]{ Seq$.\j$};
}
}};
\draw [->, thick, white!5!gray] ($(outer.west)-(-1.5ex,27.9ex)$) -- ($(outer.east)-(1.5ex,27.9ex)$) node (t0) [xshift=-111ex,yshift=-1ex]{\color{black}  $t_0$} node (t1) [xshift=0ex,yshift=-1ex]{\color{black}  $T$} node (tT) [xshift=-55ex,yshift=-1ex]{\color{black} $t_0 < T$};
}};
%
%\node (boundary) [draw=gray,rounded corners=0.4mm,minimum width=112ex,minimum height=54ex]{};
%
\end{tikzpicture}
\label{fig_tracking}
}
\subfloat[]
{
\begin{tikzpicture}
\node (outer) [scale=1]{
\tikz{
\foreach \i in {0,1,2}
{
\foreach \j in {0,1,2,3,4}
{
\node (t\j\i) [draw=none,rounded corners, xshift=\j*9.5ex, yshift=-\i*7.5ex]{\includegraphics[width=9.4ex,height=7ex]{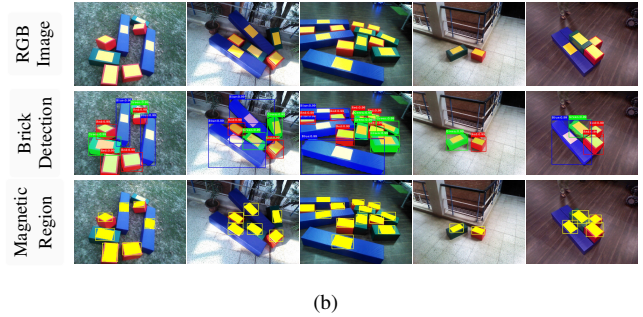}};
}
}
%
%\colorlet{fillcolor}{white!90!black}
\colorlet{fillcolor}{white!98!black}
\node (a) [draw=white,fill=fillcolor,rounded corners=0.6mm, xshift=-8ex, yshift=-0ex,rotate=90]{\shortstack{\scriptsize \scriptsize RGB \\ \scriptsize Image}};
\node (a) [draw=white,fill=fillcolor,rounded corners=0.6mm, xshift=-8ex, yshift=-7.5ex,rotate=90]{\shortstack{\scriptsize Brick \\ \scriptsize Detection}};
\node (a) [draw=white,fill=fillcolor,rounded corners=0.6mm, xshift=-8ex, yshift=-15ex,rotate=90]{\shortstack{\scriptsize Magnetic \\ \scriptsize Region}};
}};
\end{tikzpicture}
\label{fig_detection}
}
\caption{ Qualitative results for (a) Tracking, and (b) Instance detection, segmentation and part detection.}
\end{figure*}
In order to justify the validity and usefulness of our work, we provide the experimental analysis of the timing performance and accuracy of detector, tracker and grasp state classifier in the context of real-time autonomous robotic applications. Performance evaluation of the visual perception system on the public benchmarks for object detection such as MS-COCO, PASCAL-VOC is out of the scope of this paper.
\subsection{Dataset}
\subsubsection{\textbf{Detection}}
The data collection process is based on our work \cite{deepquick} which won $3$rd prize in Amazon Robotics Challenge, $2017$. Following \cite{deepquick}, we collect a very small sized dataset of about $100$ images which contains multiple instances of all four categories. We split the dataset into training and testing set in the ratio of $7:3$. We perform both box regression and segmentation for the instances while only mask segmentation is performed for the ferromagnetic regions. In order to achieve that, for each instance in an image, two masks are generated manually, one for instance itself and another for its ferromagnetic region. These masks serve as groundtruth for instance segmentation and part segmentation pipeline. The ground truth for box regression is extracted automatically based on the convex hull of the mask pixels belonging to the instance. The instance class-ID is defined at the time of mask generation. In order to multiply the dataset size, run time occlusion aware scene synthesis is performed similar to \cite{deepquick}. The synthetic scene synthesis technique allows us to annotate only a few images and appears to be a powerful tool for data multiplication.
\subsubsection{\textbf{Tracking}}
We collect $10$ indoor and $10$ outdoor video sequences at $\sim30$FPS with instance size varying from smaller to larger. We downsample the video sequences to $\sim2$ FPS and annotate each frame for instance detection, classification, segmentation and part segmentation. The overall dataset for tracking purpose roughly consists of $200$ images. The $100$ indoor images are used for training and $100$ outdoor images are kept for testing. This is done in order to examine the generalization of the tracker. Synthetic scene generation \cite{deepquick} plays an important role in this process.
\subsubsection{\textbf{Grasp State Classification}}
In order to train the grasp state classifier, we collect four kinds of images ($50$ each).
\begin{enumerate}
\item The foam in compressed state with object attached.
\item The foam in compressed state without object attached.
\item The foam in uncompressed state with object attached.
\item The foam in uncompressed state with no object.
\end{enumerate}
The sets of images $\{1,2\}$ and $\{3,4\}$ represent the states $G_1$, $G_2$, $G_3$, $G_4$. All the collected images are cropped and resized as described in Sec.\ref{subsec_graspstate}. Practically, the actual cropping region is kept slightly larger than discussed above. It is done in order to provide contextual details for better classification performance. The dataset is splitted into $2:3$ for training and testing i.e. $40$ training and $60$ test images are available for each category. While training, we again utilize synthetic scene synthesis to accommodate for the dynamic outdoor scenes. For more details, please refer to \cite{deepquick}.
\subsection{Training Policy}
We refer the baseline architecture as Arch$_1$. We perform training and testing on three different variants Arch$_1$, Arch$_2$ and Arch$_3$. The Arch$_2$ and Arch$_3$ consists of $25\%$ and $50\%$ more number of parameteres as compared to Arch$_1$, in each layer of all CNN backbones. The kernel sizes for various layers have been provided in Table-\ref{tab_kernelsize}. 
\par
Since, detection and tracking are two different tasks, therefore, separate training is required. In order to avoid that, instances and parts are also annotated along with the masks in the training dataset of the tracker. In other words, the detector now can be trained on the tracker training data. To balance the training process, images are chosen randomly with a probability of $0.5$ from both the detector and the tracker training dataset. Through this technique, the unified system experiences glimpse of temporal data (due to presence of video sequences) as well as ordinary single image object detection. 
\par
Further, we report mean-intersection-over-union (mIoU) scores for box detection, instance segmentation, and part segmentation. The mIoU score is a well known metric for reporting box regression and segmentation performance. Since, tracker essentially performs segmentation, therefore same is reported for the tracker. Due to very high inter-class and low intra-class variance, we did not encountered any misclassification of the bricks. Therefore, we have dropped the box classification accuracy from the results. 
\subsection{Training Hyperparameters}
We use base learning rate = $0.01$, learning rate policy=$step$ , $SGD$ optimizer with nestrov momentum = $0.9$ for pretraining the backbone on mini imagenet dataset. While base learning rate = $0.001$, learning rate policy=$step$, $ADAM$ optimizer with $\beta_1=0.9$ and $\beta_2=0.99$ are used for finetuning on our dataset.%
\begin{table}[t]
\centering
\caption{\footnotesize Kernel sizes. `$*$' equals to the input channels}
\label{tab_kernelsize}

\arrayrulecolor{white!60!black}

\begin{tabular}{c|c|c|c}
\hline
Layer & $C_1$ & $C_2$ & $C_3$ \\ \hline
Stage-$1$ & $4\times3\times3\times3$ & $2\times1\times3\times3$ & $2\times3\times3\times3$ \\
Stage-$2$ & $8\times4\times3\times3$ & $4\times2\times3\times3$ & $4\times2\times3\times3$ \\
Stage-$3$ & $16\times8\times3\times3$ & $4\times4\times3\times3$ & $8\times4\times3\times3$ \\
Stage-$4$ & $32\times16\times3\times3$ & $8\times4\times3\times3$ & $8\times8\times3\times3$ \\
Stage-$5$ & $32\times32\times3\times3$ & $16\times8\times3\times3$ & $16\times8\times3\times3$ \\
Others & $12\times*\times3\times3$ & $4\times*\times3\times3$ & $8\times*\times3\times3$ \\ \hline
\end{tabular}
\end{table}
\begin{table*}[t]

\centering
\caption{\footnotesize Effect of Synthetic Scenes and Comprehensive Data Augmentation fro Arch$_1$, FP$16$}
\label{tab_augmentation}

\arrayrulecolor{white!60!black}

\begin{tabular}{c c c c c c c c c c}
\hline

\multicolumn{6}{c}{Augmentation} & \multirow{2}{*}{\makecell{Box Detection \\ mIoU}} & \multirow{2}{*}{\makecell{Segmentation\\ mIoU}} & \multirow{2}{*}{\makecell{Part Segmentation\\ mIoU}} & \multirow{2}{*}{\makecell{Tracker \\ mIoU}}\\ \cline{1-6}

colour & scale & mirror & blur & rotate & synthetic scenes & \\ \hline
 \xmark  & \xmark & \xmark & \xmark & \xmark & \xmark & \multicolumn{1}{|c}{$23.7$} & $31.8$ & $18.8$ & $15.3$\\ 
 \cmark  &  &  &  &  & & \multicolumn{1}{|c}{$25.3$} & $32.1$ & $21.3$ & $17.7$ \\ 
 \cmark  & \cmark &  &  &  &  &\multicolumn{1}{|c}{$30.1$} & $38.6$& $29.1$& $23.1$\\ 
 \cmark  & \cmark & \cmark &  &  &  &\multicolumn{1}{|c}{$32.3$} & $40.2$& $31.9$ & $25.5$\\ 
 \cmark  & \cmark & \cmark & \cmark &  &  &\multicolumn{1}{|c}{$32.5$} & $41.4$& $33.8$ & \textcolor{red}{$24.1$}\\ 
 \cmark  & \cmark & \cmark & \cmark & \cmark &  &\multicolumn{1}{|c}{$37.2$} & $49.8$ & $37.7$ & $28.4$\\ %\hline
 \cmark  & \cmark & \cmark & \cmark & \cmark & \cmark & \multicolumn{1}{|c}{\textcolor{blue}{$80.3$}} & \textcolor{blue}{$79.2$} & \textcolor{blue}{$73.1$} & \textcolor{blue}{$83.4$}\\ \hline

\end{tabular}
\end{table*}
\subsection{Ablation Study of Perception System}
Table-\ref{tab_performance} depicts the performance of various perception components along with the timing performance. For each performance metric, we perform ablation study of all the three architectures Arch$_1$, Arch$_2$, and Arch$_3$. It can be noticed that Arch$_2$ shows an minor improvement over Arch$_1$, however, Arch$_3$ does not show significant improvements despite an increment in the number of learnable parameters by $50\%$. In our views, It happens because the backbone architecture for all the three variants remains same except the representation power. Another reason for that is the objects under consideration are of uniform colours and therefore, intra-class variance is very low. 
\par
The table also shows various performance metrics for both Half-Precision (FP$16$) and single-precision (FP$32$) computations on GPU. A single precision floating point number consists of $32$ bits whereas half precision consists of $16$ bits. There also exists double precision FP$64$ which consists of $64$ bits. As the number of bits of a floating point number increases, the smallest number representable decreases i.e. more details comes in. However, the amount of computation time required increases non-linearly and drastically. Therefore, in practice FP$32$ is used on desktop grade GPUs. However, due to real time constraints and power considerations, NVIDIA has recently developed half-precision based high-performance libraries especially for deep learning algorithms. We take advantage of half precision in our application without sacrificing the accuracy.
\begin{table}[t]
\centering
\caption{\footnotesize Performance Analysis of Perception System}
\label{tab_performance}

\arrayrulecolor{white!60!black}

\begin{tabular}{c|c|c|c|c|c}
\hline
\makecell{Network \\ Architecture} & \makecell{Time \\ (ms)} & \makecell{Box \\ mIoU} & \makecell{Seg \\ mIoU} & \makecell{Part Seg \\ mIoU} & \makecell{Tracker \\ mIoU}  \\ \hline
 Arch$_1$ (FP$16$) & \textcolor{blue}{$35$} & $80.3$  & $79.2$  & $73.1$ & $83.4$ \\ %\hline
 Arch$_1$ (FP$32$) & $69$ & $82.2$  & $80.0$  & $72.9$ & $84.8$ \\ \hline
 Arch$_2$ (FP$16$) & $41$ & $81.2$  & $79.7.$ & $73.6$ & $84.1$ \\ %\hline
 Arch$_2$ (FP$32$) & $73$ & $83.5$  & $81.1$  & $74.0$ & $85.6$ \\ \hline
 Arch$_3$ (FP$16$) & $52$ & $85.3$  & $80.6$  & $73.5$ & $84.4$ \\ %\hline
 Arch$_3$ (FP$32$) & $97$ & \textcolor{blue}{$86.7$}  & \textcolor{blue}{$81.9$}  & \textcolor{blue}{$74.1$} & \textcolor{blue}{$86.4$} \\ \hline

\end{tabular}
\end{table}
\subsection{Synthetic Scenes and Comprehensive Data Augmentation}
Tabel-\ref{tab_augmentation} shows the effect of synthetic scenes and other augmentations as performed in \cite{deepquick}. It can be noticed that synthetic scenes alone contribute to the improved performance as compared to the other remaining augmentations (highlighted in \textcolor{blue}{Blue}). The decreasing value of tracker mIoU is observed when blur is included in the augmentation process.
\subsection{Unified vs Distributed Perception System}
The unification of various tasks is a primary contribution of this paper. Hence, we compare the timing performance of the unified system against three different networks for detection, part segmentation and tracking (Table-\ref{tab_unified}). These isolated networks have exactly same configurations for each layer in CNNs and other required components such as RPN, FPN etc. It is clearly evident from the table that our unified pipeline is far better in terms of timing performance, because all the three modules can be run simultaneously. Fig. \ref{fig_detection} and Fig. \ref{fig_tracking} show qualitative results of unified perception system. The results in Table-\ref{tab_unified} are with Arch$_1$ trained using all kinds of augmentations.
\subsection{Grasp State Classification}
The performance of grasp state feedback CNN is shown in Table-\ref{tab_graspstate}. The definitions of Arch$_1$, Arch$_2$, and Arch$_3$ remains same, except the parameters of Arch$_1$ belong to the Grasp State Classifier $C_3$ (Fig. \ref{fig_cnn}). The $100\%$ accuracy is evident due to very simple classification task. Also, it is important to note that, for both Arch$_1$ and Arch$_2$, the timing performance of FP$16$ is roughly twice of FP$32$ where as it is not the case with Arch$_3$ (highlighted in red). It happens because of the non-linear and saturating speed-up curves of the GPU device.
\begin{table}[t]
\centering
\caption{\footnotesize Unified Vs Distributed Perception System, Arch$_1$, FP$16$}
\label{tab_unified}

\arrayrulecolor{white!60!black}

\begin{tabular}{c|c|c|c|c}
\hline
\multirow{2}{*}{\makecell{Network \\ Architecture}} & \multicolumn{4}{c}{Time (ms)} \\ \cline{2-5}

 & \makecell{Detection} & \makecell{Part Seg} & \makecell{Tracking}   & \makecell{Total} \\ \hline
 Arch$_1$ &  $--$  & $--$  & $--$ & \textcolor{blue}{$35$} \\ \hline
 Detection & $29$  & $--$  & $--$ & \multirow{2}{*}{$92$} \\ %\hline
 Part Segmentation & $--$  & $30$  & $--$ & \\ %\hline
 Tracking & $--$ & $--$  & $33$ & \\ \hline

\end{tabular}
\end{table}

\begin{table}[t]
\centering
\caption{\footnotesize Performance Analysis of Grasp State Classification}
\label{tab_graspstate}

\arrayrulecolor{white!60!black}

\begin{tabular}{c|c|c|c}
\hline
\makecell{Network \\ Architecture} & \makecell{Time \\ (ms)} & \makecell{Foam State\\ Accuracy($\%$)}  & \makecell{Object Attached State\\ Accuracy($\%$)}  \\ \hline
 Arch$_1$ (FP$16$) & $15$ & $100.0$ & $100.0$\\ %\hline
 Arch$_1$ (FP$32$) & $32$ & $100.0$ & $100.0$\\ \hline
 Arch$_2$ (FP$16$) & $20$ & $100.0$ & $100.0$\\ %\hline
 Arch$_2$ (FP$32$) & $41$ & $100.0$ & $100.0$\\ \hline
 Arch$_3$ (FP$16$) & $25$ & $100.0$ & $100.0$\\ %\hline
 Arch$_3$ (FP$32$) & \text{\color{red}$56$} & $100.0$ & $100.0$\\ \hline

\end{tabular}
\end{table}

\section{Conclusion}
\label{sec_conc}
This work introduces an end-to-end framework for UAV based manipulations tasks in GPS-denied environments. A deep learning based real-time unified visual perception system is developed which combines the primary tasks of instance detection, segmentation, part segmentation and object tracking. The perception system can run at $30$ FPS on Jetson TX$2$ board. A complete electromagnets based gripper design is proposed. A novel approach to handle the grasp state feedback is also developed in order to avoid external electronics. To the best of our knowledge, the unified vision system combining four tasks altogether and the grasp state feedback in the context of UAVs has not been developed in the literature. Apart from that, a remote computing based approach for 6DoF state-estimation is introduced. Certain improvements in the opensource framework MoveIt! to accommodate UAVs have also been done.

%\nocite{*}
\bibliographystyle{ieeetr}
\bibliography{bibfile}

\end{document}